\documentclass[journal]{IEEEtran}
\usepackage{cite}

\ifCLASSINFOpdf
\else
\fi
\usepackage{graphicx}
\graphicspath{{img/}}
\DeclareGraphicsExtensions{{.eps},{.pdf}}
\usepackage{epsfig}
\usepackage{subfigure}
\usepackage{color}

\usepackage{amsmath}
\usepackage{amsfonts}
\usepackage{textcomp}
\usepackage{algorithm}
\usepackage{multirow}
\usepackage{caption}	

\usepackage[normalem]{ulem}
\usepackage{xstring}

\usepackage[OT1]{fontenc} 

\usepackage{algpseudocode}

\usepackage{dblfloatfix}
\usepackage{url}

\newcommand{\remove}[2][1]{\IfEqCase{#1}{%
		{0}{{\color[rgb]{1,0,0}\sout{#2}}}
		{1}{}
	}
	[\PackageError{action}{Undefined option tor action:#1}{}]
}
\newcommand{\myremove}[1]{\remove[1]{#1}} 
\newcommand{\add}[2][1]{\IfEqCase{#1}{%
		{0}{{\color[rgb]{0,0,1}\uline{#2}}}
		{1}{{#2}}
	}
	[\PackageError{action}{Undefined option tor action:#1}{}]
}
\newcommand{\myadd}[1]{\add[1]{#1}} 

\begin{document}
%
\title{Synchronous Maneuver Searching and Trajectory \\Planning for Autonomous Vehicles in Dynamic Traffic Environments}
%
%
%

\author{Lilin Qian, Xin Xu, Yujun Zeng, Xiaohui Li, Zhenping Sun, and Hang Song 
\thanks{*This work was supported by National Natural Science Foundation of China under Grant 61751311, 61825305, U1564214.}
\thanks{Lilin Qian, Xin Xu, Yujun Zeng, Xiaohui Li, Zengping Sun and Hang Song are with College of Intelligence Science and Technology, NUDT, Changsha, China.
}
}

\maketitle


\begin{abstract}
\myremove{For}\myadd{In}the real-time decision-making and local planning process of autonomous vehicles in dynamic environments, the autonomous driving system may fail to find a reasonable policy or even gets trapped in some situation due to the complexity of global tasks and the incompatibility between \myadd{upper level}maneuver decisions with the\myremove{low-level}\myadd{lower level}trajectory planning. To solve this problem, this paper presents 
a synchronous maneuver searching and trajectory planning (SMSTP) algorithm based on the topological concept of homotopy. Firstly, a set of alternative maneuvers with boundary limits are enumerated on a multi-lane road. Instead of sampling numerous paths in the whole spatio-temporal space, we, for the first time, propose using Trajectory Profiles (TPs) to quickly construct the topological maneuvers represented by different routes, and put forward a corridor generation algorithm based on graph-search. The bounded corridor further constrains the maneuver's space in the spatial space.
A step-wise heuristic optimization algorithm is then proposed to synchronously generate a feasible trajectory for each maneuver. To achieve real-time performance, we initialize the states to be optimized with the boundary constraints of maneuvers, and we set some heuristic states as terminal targets in the quadratic cost function.  The solution of a feasible trajectory is always guaranteed only if a specific maneuver is given. The simulation and realistic driving-test experiments verified that the proposed SMSTP algorithm has a short computation time which is less than 37ms, and the experimental results showed the validity and effectiveness of the SMSTP algorithm.
\end{abstract}


%
\IEEEpeerreviewmaketitle

\section{Introduction}
%
%
%
%
\IEEEPARstart{A}{utonomous} driving has been extensively studied in the past three decades and has a wide variety of applications, especially in intelligent transportation systems. However, the thriving of autonomous driving does not give birth to a sufficiently-developed pilot system. There are still many research challenges in developing autonomous driving systems in complex environments. Among these challenges, real-time maneuver decision and local planning are two key technologies for dealing with dynamic traffic. Generally, a typical hierarchical maneuver reasoning and trajectory planning system is applied in most autonomous vehicles \cite{buehler2009DARPAUrbanChallengea}. The hierarchical design can result in inconsistent situations. The upper\myremove{layer}\myadd{level}of maneuver reasoning does not guarantee a favourable space for the lower\myremove{layer}\myadd{level}of trajectory planning. Besides, maneuver reasoning in only spatial space or in spatio-temporal space with a short horizon does not discover more reasonable maneuvers, especially in the scenarios of lane merging, turning at intersections, etc.
In the maneuver decision\myremove{layer}\myadd{level,}various drive strategies are decided according to global task (e.g. speed limit, intersection precedence handling), and local traffic (e.g. surrounding vehicles, obstacles in lane). Traditionally, the rule-based finite state machine  \cite{montemerlo2009JuniorStanfordEntry} and decision trees \cite{Miller2008Team} have been widely used for \myadd{the}maneuver decision. The experience-based rules can reliably deal with deterministic and rather simple traffic situations well, but it generalizes poorly to unknown situations. The problem lies in that the artificial rules is not robust and completeness are not guaranteed. At the same time, machine learning based decision methods also have been investigated elaborately {\cite{Bahram2016AGame,lenz2016TacticalCooperativePlanning, li2017GameTheoreticModeling,talebpourModelingLaneChangingBehavior2015}}. These methods model the interactions between the agent and other traffic participants in discrete actions and search an optimal action in a tree graph. Although these methods have developed traditional rules in some specific traffic situations, they still rely on artificially designed states or hand-crafted feature sets, and sometimes have problems with oscillatory behaviours and integrate the traffic rules poorly. 

For the  lower\myremove{layer}\myadd{level}of trajectory planning, given a certain maneuver, it searches a set of executable trajectories connecting the start state space to target state space independently\cite{johnson2012Optimal}. David et al. \cite{gonzalez2015A} have surveyed several methods of trajectory planners for autonomous vehicles, which will not be repeated here. A hierarchical pilot system may fail in tough situations, since the conservative maneuver decision can hardly guarantee enough space to plan a safe and comfort trajectory. A reasonable pilot system requires that the trajectory planner should distinguish between maneuvers and the maneuver-decision maker should guarantee a feasible trajectory at the same time.

\begin{figure*}[!t]
	\centering
	\includegraphics[width=0.95\linewidth,height=4cm]{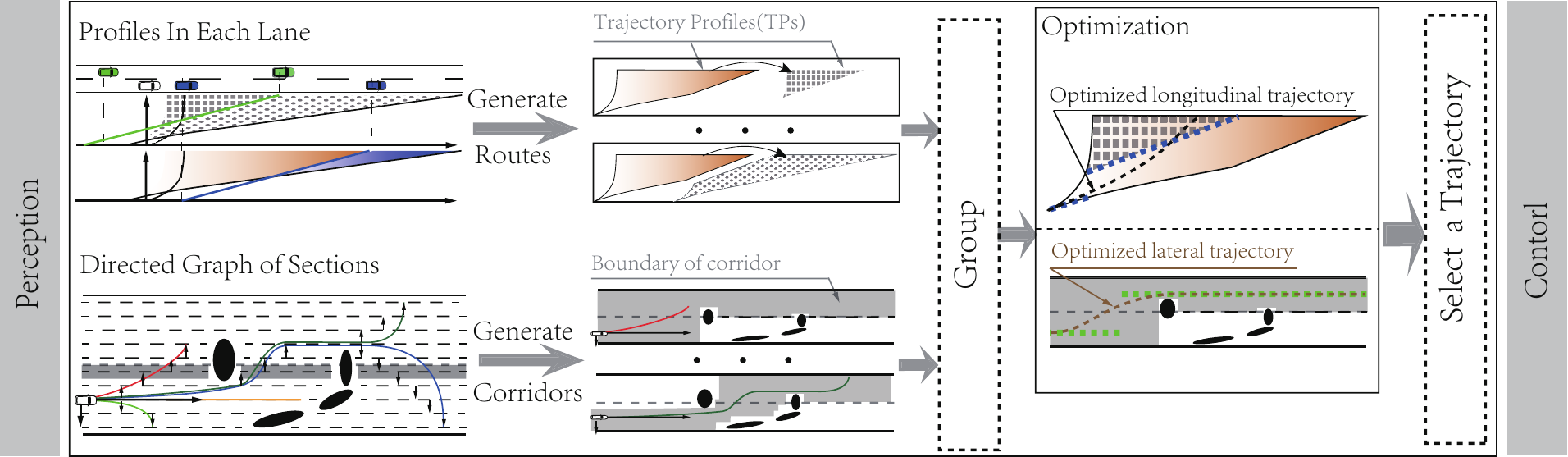}
	\caption{System architecture of SMSTP algorithm.}
	\label{fig:sys}
\end{figure*}

Apparently, a close integration of \myadd{the}maneuver decision and trajectory planning requires that all feasible tunnels should be extracted from the whole spatio-temporal space. Similarly, human drivers find a desired tunnel with a quick glance at the surrounding environments. Chen \cite{chen2005TopologicalApproachPerceptual} indicates that the global topological perception is prior to the perception of other geometrical properties. In other words, the topological perception based on physical connectivity emerges earlier than the geometric perception. The topological perception occurs in only five milliseconds at the early stages of visual perception, and this explains why human drivers are talented in making decision\myadd{s.}Taking the lane change, for example, a lane change can be divided into mandatory lane change (occurs when a vehicle must get into the target lane in limited time or space) and  discretionary lane change (maintains high cruise speed in natural traffic flow). Obviously, a lane change involves a dichotomy between the endogenously generated signals (e.g. traffic rules, drive styles or the motivation) and physical stimuli (e.g. lane boundaries, static obstacles and dynamic objects)\cite{bisley2010AttentionIntentionPriority}. The combined dichotomy determines task-level expectation of a driver whilst the physical stimuli decide the topological perception that is, the tunnels in the spatio-temporal space. Inspired by this, we mainly focus on maneuver searching based on the topological tunnels in this paper.

Currently, two main topology-based methods have been extensively investigated, namely, the sampling based methods and the combinatorial methods. Sampling based methods group paths laterally shifted from a lane-center with motion primitives smoothly connected \cite{gu2016AutomatedTacticalManeuver,gu2017ImprovedTrajectoryPlanning,sontges2018ComputingDrivableArea}. The maneuvers are\myremove{defined}\myadd{generated}by clustering paths into the same groups or homotopy classes. The discrete sampling sacrifices optimality and the fixed look-ahead time will prevent the algorithm from finding more smoother maneuvers. In addition, sample-based methods have problems \myadd{in}dealing traffic rules where precedence or other semantic elements must be considered. The second kind is the combinatorial method using a divide-and-conquer strategy. Firstly, different convex decomposition methods, such as cell decomposition and reference-frame methods are utilized to enumerate different homotopy classes in the unstructured environments  \cite{Bes2012Path, bhattacharya2012TopologicalConstraintsSearchbased, bhattacharya2015PersistentHomologyPath,kuderer2014OnlineGenerationHomotopicallya,rosmann2017IntegratedOnlineTrajectorya}, then suboptimal paths are generated in each homotopy class. The idea of generating distinct topologies first of all, has also been used for on-road situation \cite{zhan2017SpatiallypartitionedEnvironmental, park2015HomotopyBasedDivideandConquerStrategy,bender2015CombinatorialAspectMotiona}. Bender et al.\cite{bender2015CombinatorialAspectMotiona} distinguish different maneuvers according to surrounding dynamic vehicles. However, these methods are merely applicable to quasi-static traffic situations. Florent et al.\cite{altche2017PartitioningFreeSpacetime} extend the planning problem into 3D space by partitioning the collision-free space in discrete time, which results in a graph method for a deep path search with only one path generated. Then, the trajectory generation problem is solved in decomposed non-convex space by using quadratic optimizing approach. Moreover, the optimization sometimes finishes without a feasible solution. In contrast to the proposed SMSTP algorithm, these methods have two main disadvantages in dealing with on-road traffic. 
To the authors\textquotesingle s knowledge, the first disadvantage is that current methods aim at finding out only one optimal trajectory either by sampling in the spatio-temporal space, or by solving a non-convex optimization problem. Even if these methods find only one optimal solution, they require either heavy computation load or excessive time for a real-time system respectively. The other is that these methods do focus on the optimal trajectory. However, to deal with the complex traffic, the expectation and preference of passengers vary at different situations thus leading to a discrete changing of the cost function. Only one optimal solution to a specific cost function cannot deal with situations where multiple alternative policies are needed. 

The main difficulty of optimization-based method lies in that generating a good initialization of the solution is never easy. Apart from that, focusing on a global optimal trajectory tends to face a dilemma in many circumstances. The optimum refers to the best solution under given conditions with respect to a quadratic cost function. Nevertheless, the maneuver searching and trajectory planning problem for autonomous vehicles cannot be represented in a single cost function. The reason is that the maneuver decision is discrete, and the cost function varies in different traffic environments and long short-term expectations. In addition, optimum requires a huge cost for computational time and space for a real-time system. Nevertheless, sometimes a solution is not guaranteed. As a matter of fact, a feasible and comfort trajectory satisfies the expectation of most people in regular driving circumstances. Thus, instead of searching a global optimal trajectory, we focus on finding out a group of alternative solutions in different maneuvers each of which has a smooth trajectory. 

In this work, a novel synchronous maneuver searching and trajectory planning (SMSTP) algorithm is proposed based on the topological concept of homotopy, whose overall architecture is shown in Fig.\ref{fig:sys} The key contributions  are two-folds. Firstly, we propose a topological maneuver searching method by partitioning the spatio-temporal space\myremove{in}\myadd{into}two 2D planes and reassembling the matched corridors and routes in each plane. We come up with TPs in adjacent $ s $\nobreakdash--$ t $ planes for the first time to represent a topological route. A TP is a profile combining different segments of vehicles\textquotesingle\ trajectories in the $ s $\nobreakdash--$ t $ plane. A TP reveals a compact space that a series of trajectory points can locate in. By using TPs to represent a homotopy route, SMSTP avoids sampling discretely in the temporal dimension and gets a closed space for path planning without any collision area inside. In addition, we propose an effective corridor generation algorithm by searching a tree structure. The corridors constrain the lateral width of the maneuvers in the $ d $\nobreakdash--$ s $ plane. Different from traditional methods using fixed sampling distance, our method faithfully splits the bands laterally shifted from the lane-center into sections according to the distribution of obstacles. Finally, the combination of routes and corridors\myremove{results in the topological maneuver}\myadd{gives birth to the topological maneuvers.}

Secondly, we present a step-wise numerical optimization method to generate a smooth trajectory. The algorithm uses adjusted weights of cost\myremove{term}\myadd{functions}to balance the solution and constraints for longitudinal and lateral\myremove{optimization}\myadd{optimizations}successively, where the boundary constraints of a maneuver are given as initial values and a heuristic state is set as the terminal target while optimizing. Thus, the massive computational time cost for non-convex optimization is avoided. The proposed algorithm is validated in simulation and real traffic. The realistic experimental test results show the rapidity and adaptability of the SMSTP algorithm to various dynamic traffic.

\myadd{Note that the SMSTP algorithm cannot be directly used in urban traffic. The complex intersections and sharp turns require a whole picture of surrounding roads, thus the HD (High definition) map and the road topology are required. Given these prerequisites, the SMSTP algorithm can be used in complex environments.}

The remainder of this paper is organized as follows. Section \ref{sec_problem_definition} introduces basic problem definitions. Section \ref{sec_topology_generation} presents the algorithm of generating different topological corridors and routes in detail. Section \ref{sec_trajectory_optimizaiont} describes the numerical optimization method for\myremove{our}trajectory generation. Section \ref{sec_exp} gives the experiments of both simulation and driving test results to evaluate the proposed algorithm. Finally, Section \ref{sec_conclusion} concludes this paper.


\section{Problem Definition}
\label{sec_problem_definition}
\subsection{Notations}

\begin{figure}[htbp]
	\centering
	\subfigure[\myadd{Three orthographic views of the trajectories in Fig.}\ref{homotopy:static-whole-lane}]{
		\begin{minipage}[t]{1\linewidth}
			\centering
			\includegraphics[width=0.98\columnwidth,height=0.3\columnwidth]{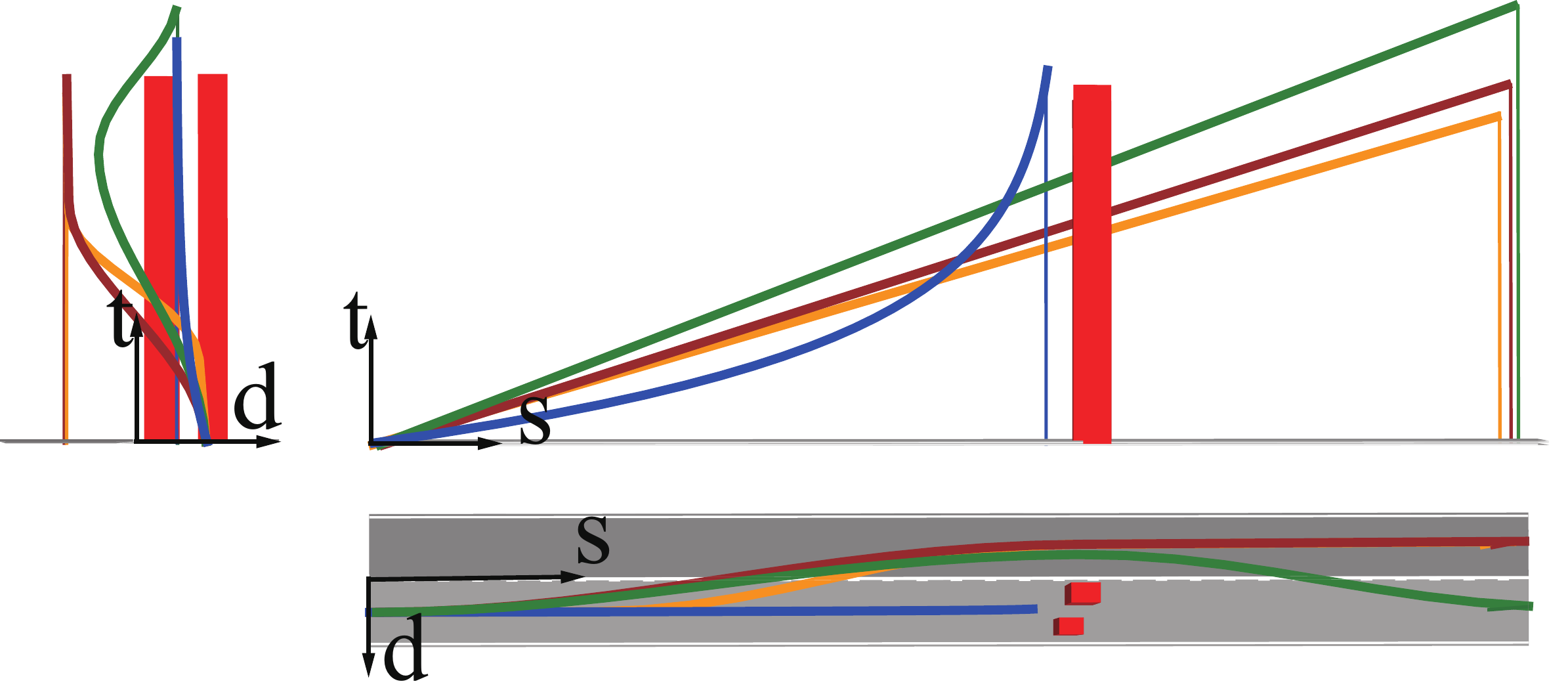}
			\label{homotopy:static-3view}
		\end{minipage}}
		
	\subfigure[\myadd{Global view of right lane totally blocked. Three homotopy paths exist.}]{
		\begin{minipage}[t]{1\linewidth}
			\centering
			\includegraphics[width=0.95\columnwidth,height=0.3\columnwidth]{fig2a}
			\label{homotopy:static-whole-lane}
	\end{minipage}}
	
	\subfigure[\myadd{Global view of right lane partly blocked. Five homotopy paths exist.}]{
		\begin{minipage}[t]{1\linewidth}
			\centering
			\includegraphics[width=0.95\columnwidth,height=0.3\columnwidth]{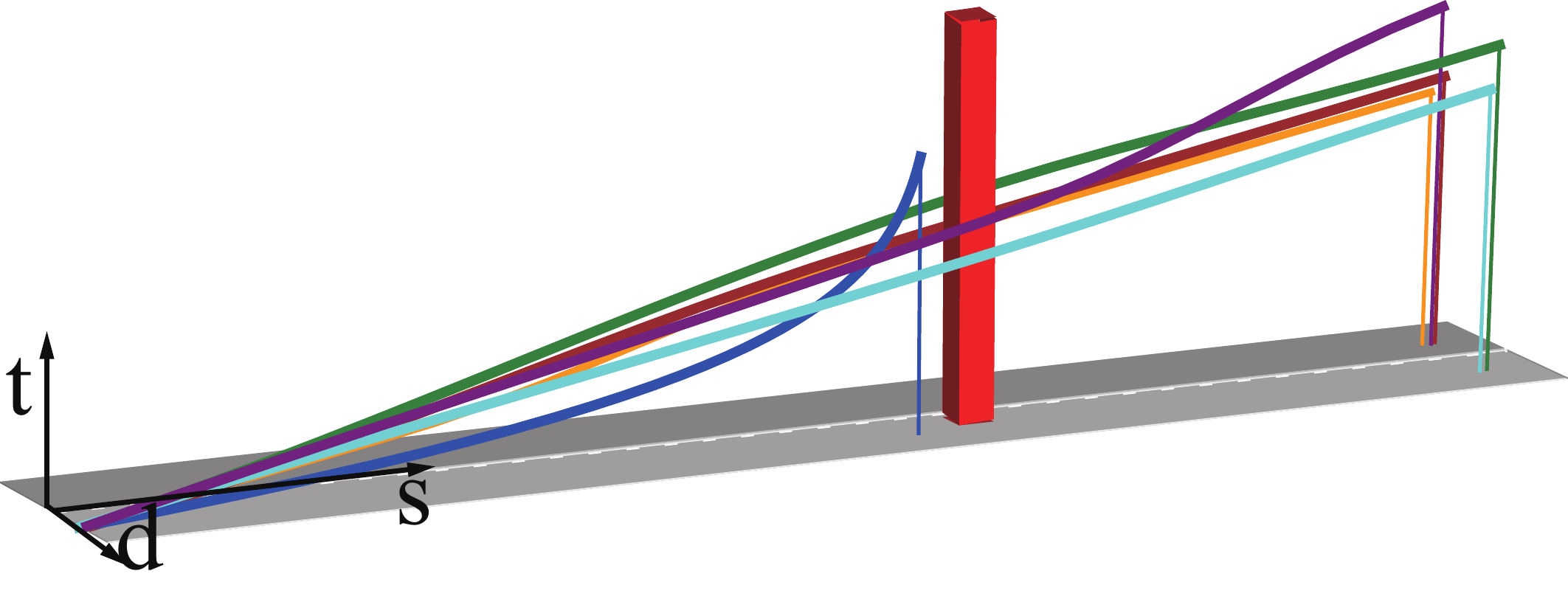}
			\label{homotopy:static-part-lane}
	\end{minipage}}
	\subfigure[\myadd{Homotopy paths in traffic flow(boxes with the same gradient colors represent the same vehicle at different times).}]{\begin{minipage}[t]{1\linewidth}
		\centering
		\includegraphics[width=0.9\columnwidth,height=0.45\columnwidth]{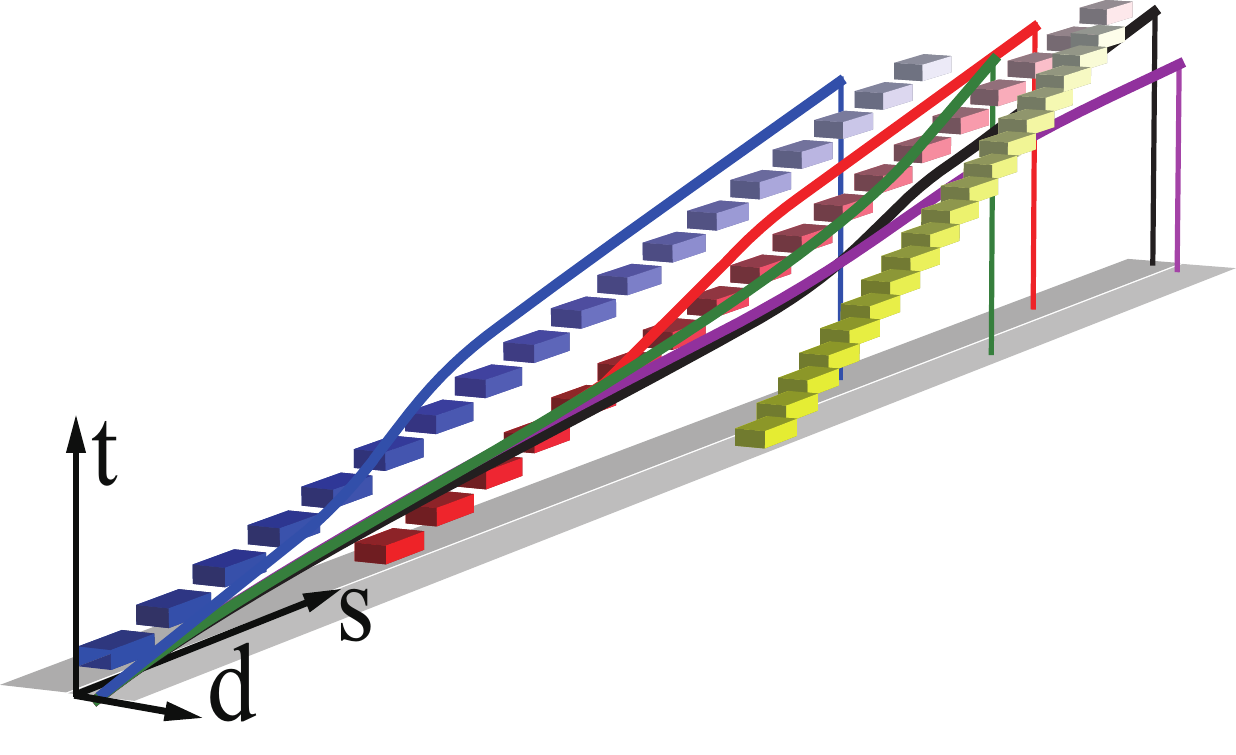}
		\label{homotopy:dynamic-vehicles}
	\end{minipage}}
	
	\caption{\myadd{Homotopy paths in different scenarios. The long red boxes represent static obstacles. The orange and yellow trajectories belong to the same homotopy path. The trajectories in other colors represent different homotopy paths separately. $d$\nobreakdash--$s$ is the Curvilinear coordinate. $ s $\nobreakdash--axis is tangent to the lane line, $ d $\nobreakdash--axis is orthogonal to $ s $\nobreakdash--axis. $ t $\nobreakdash--axis representing time is orthogonal to $d$\nobreakdash--$s$ plane.}}
	\label{homotopy:aviable}
\end{figure}

%
%

\subsubsection{Planning Space}
As shown in Fig.\ref{homotopy:aviable}, the lane lines are boundary constraints on a vehicle, and the spaces occupied by obstacles and vehicles are the collision\myremove{-free}areas. The obstacles, the vehicles and the ego agent refer to the stationary entities, the moving traffic participants and the agent (ego autonomous vehicle) respectively in the following context without specific description. The obstacles will occupy the lane for all time. A vehicle in each lane covers either the whole lane (keeping in the lane) or two lanes (overtaking others, \myadd{changing lane}), and a vehicle moves along its trajectory in the 3D space (see Fig.\ref{homotopy:dynamic-vehicles}). The number of the obstacles can be enormous. For the convenience of computation, a cost-map, $ m_{cost} $ is generated from the obstacles. The side-ward boundaries are limited by lane lines $ L_{lines} $. 
The state of a vehicle or the agent at time $ t $ is given as $ \mathbf{x}_{i}^{t}= [s_{i}^{t}, d_{i}^{t}, v_{i}^{t}]^{\top} $, which consists of the longitudinal and lateral positions $ s_{i}^{t} $ and $ d_{i}^{t} $ in the curvilinear-coordinate \cite{chu2012LocalPathPlanninga} and the velocity $ v_{i}^{t} $.  $ \mathbf{x}_{i}=[x_{0},...,x_{T}]^{\top} $ is the trajectory of a vehicle $ i $ and $ \mathbf{x}_{e} $ is the planned trajectory of the ego agent. For simplicity, $ \mathbf{X}=\{x_{1},\dots, x_{M} \} $ is the set of trajectories with all other traffic participants, and $ M $ is the number of vehicles.

\subsubsection{Corrdinates Conversion}
\begin{figure}[!ht]
	\centering
	\includegraphics[width=0.9\columnwidth]{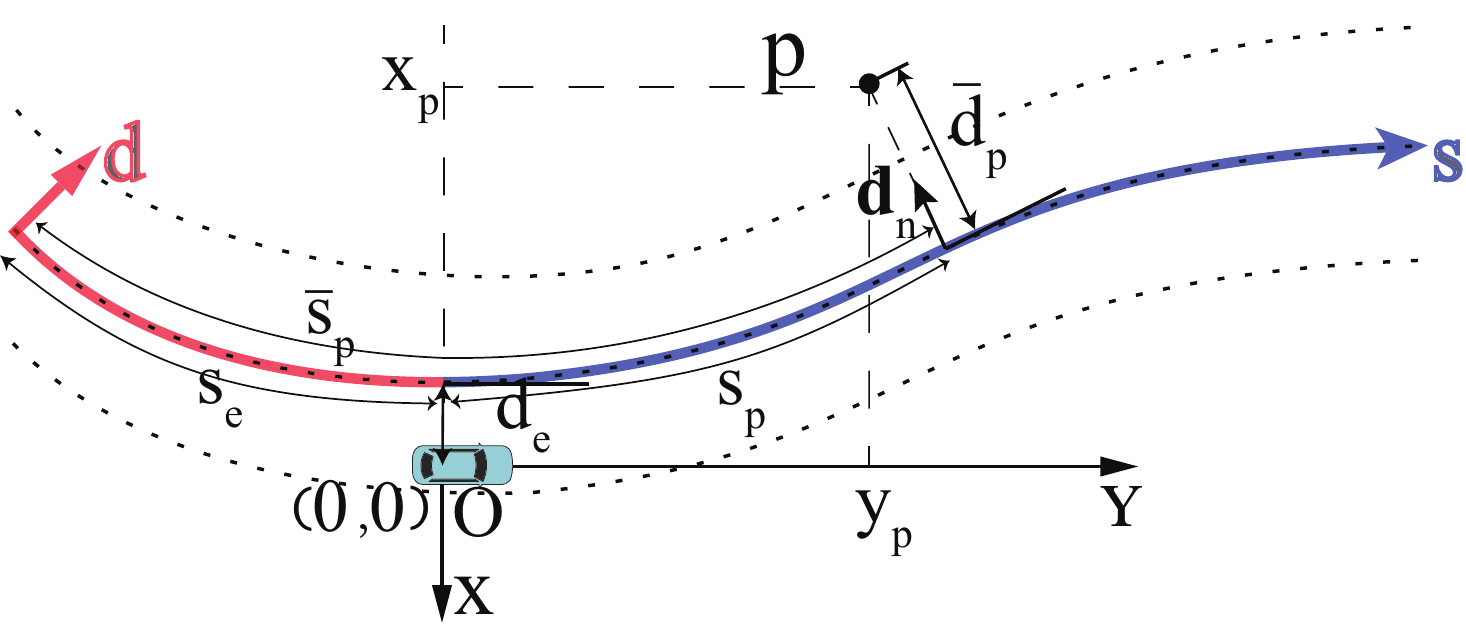}
	\caption{\myadd{Conversions between two coordinates. The colored curve is the Curvilinear coordinate. \textbf{XOY} is the Cartesian coordinate of the ego vehicle.}}
	\label{coordinate-conversion}
\end{figure}

As the lane lines are not always straight, various shapes of lane make it difficult to model lane structure in a unified form. Traditional methods use polynomial curves, splines, parabolic curves, etc \cite{narote2018ReviewRecentAdvances} to fit a line, where the resultant curves are only precise within a short range and various models must be used to cover different road shapes. Instead of fitting curves, the lanes are represented by a line that is in a set of points with fixed interval. An accumulated reference path is used to convert between Curvilinear coordinate and Cartesian coordinate as shown in Fig.\ref{coordinate-conversion}. The reference path accumulates along a lane line. The reference path includes the pose $ (x, y) $ in Cartesian coordinate, the perpendicular vector $ \mathbf{d}_{n} $, the accumulated distance $ s $ in Curvilinear coordinate and the curvature $ \rho $ of each point. For conversion \myadd{of any point $ P $}from $ (x_{p}, y_{p}) $ to $ (\bar{s}_{p}, \bar{d}_{p}) $, it searches the closest point on the reference path by binary search(this is guaranteed by a fixed interval of points in reference path)\myremove{.}\myadd{,}and the accumulated distance is $ \bar{s}_{p}$\myremove{,}\myadd{.}\myremove{the}\myadd{The}lateral shift $ \bar{d}_{p} $ is calculated by the point-to-line distance. For conversion \myadd{of the point $ P $}from $ (\bar{s}_{p}, \bar{d}_{p}) $ to $ (x_{p}, y_{p}) $, it searches the accumulated length of $ \bar{s}_{p} $ by binary search in reference path, and a shift \myadd{$ \bar{d}_p $}from the corresponding point in $ \mathbf{d}_{n} $ gets $ (x_{p}, y_{p}) $. Apparently, the pose of\myremove{the agent}\myadd{ego vehicle}is\myremove{$ (s_{ego}, d_{ego}) $}\myadd{$ (s_{e}, d_{e}) $}in Curvilinear coordinate and is $ (0, 0) $ in local Cartesian coordinate.\myremove{And}\myadd{For}conversions of other points related to vehicle, it only needs a shift to the coordinate of the\myremove{agent}\myadd{vehicle\textquotesingle s}pose.

\subsection{Homotopy Class}
\label{homotopy-decsribtion}
The utility of homotopy classes in vehicle navigation has been studied in \cite{Bes2012Path}. 
A homotopy path class is defined as a set of paths that connect the start state and the terminal state in the same topology.
Inspired by Bhattacharya\textquotesingle s work \cite{bhattacharya2012TopologicalConstraintsSearchbased}, Gu \cite{gu2016AutomatedTacticalManeuver} and Schulz \cite{schulz2017EstimationCollectiveManeuvers} extended the co-terminal-guaranteed paths to spatio-temporal trajectory planning based on the idea of pseudo-homology where co-terminal is replaced by a co-region. Relaxing some specific end states to terminal regions has also been used in \cite{zhan2017SpatiallypartitionedEnvironmental,altche2017PartitioningFreeSpacetime}. In this paper, lanes are distinguished as different topologies vividly, so the behavioral discovery is consistent with the road structure, i.e. the lane branches. Therefore, the top-level semantic instructions can be easily integrate into\myremove{a}\myadd{the}maneuver decision.

For \myadd{the}maneuver decision, lanes should be considered in a discrete manner. As show in Fig.\ref{homotopy:static-whole-lane}, three homotopy paths exist when right (forward direction) lane is totally blocked by red obstacles. The brown and orange paths going to the end of the left lane are homotopic. The blue path goes straight and stops in front of obstacles. The green path leads the agent to the end of original lane with two lane-change avoiding the obstacles. When obstacles occupy part of the right lane in Fig.\ref{homotopy:static-part-lane}, two more homotopy paths exist. One purple path goes to left lane after passing the obstacle on \myadd{the}right side. The other homotopy path leads the agent to the end of lane on \myadd{the}right \myadd{side.}Paths derived from the discretization of lanes now are corresponding to different maneuvers topologically.

\begin{figure}[htbp]
	\centering
	\subfigure[\myadd{The homotopy classes in global view.}]{\begin{minipage}[c]{1.\linewidth}
			\includegraphics[width=0.9\columnwidth, height=0.35\columnwidth]{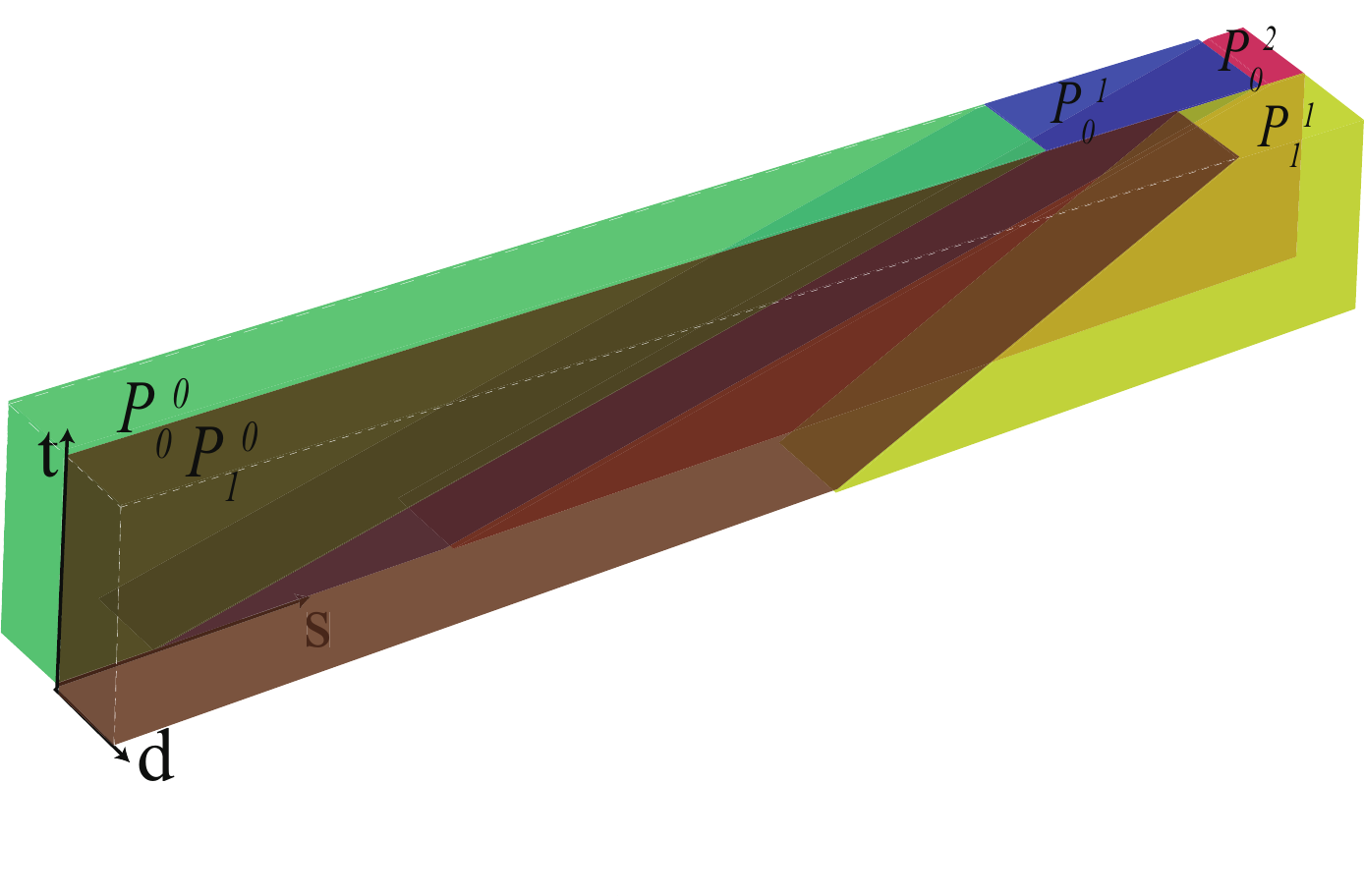}
	\end{minipage}}
	\subfigure[\myadd{Lane keep.}]{	\begin{minipage}[b]{0.3\linewidth}
			\includegraphics[width=1\columnwidth]{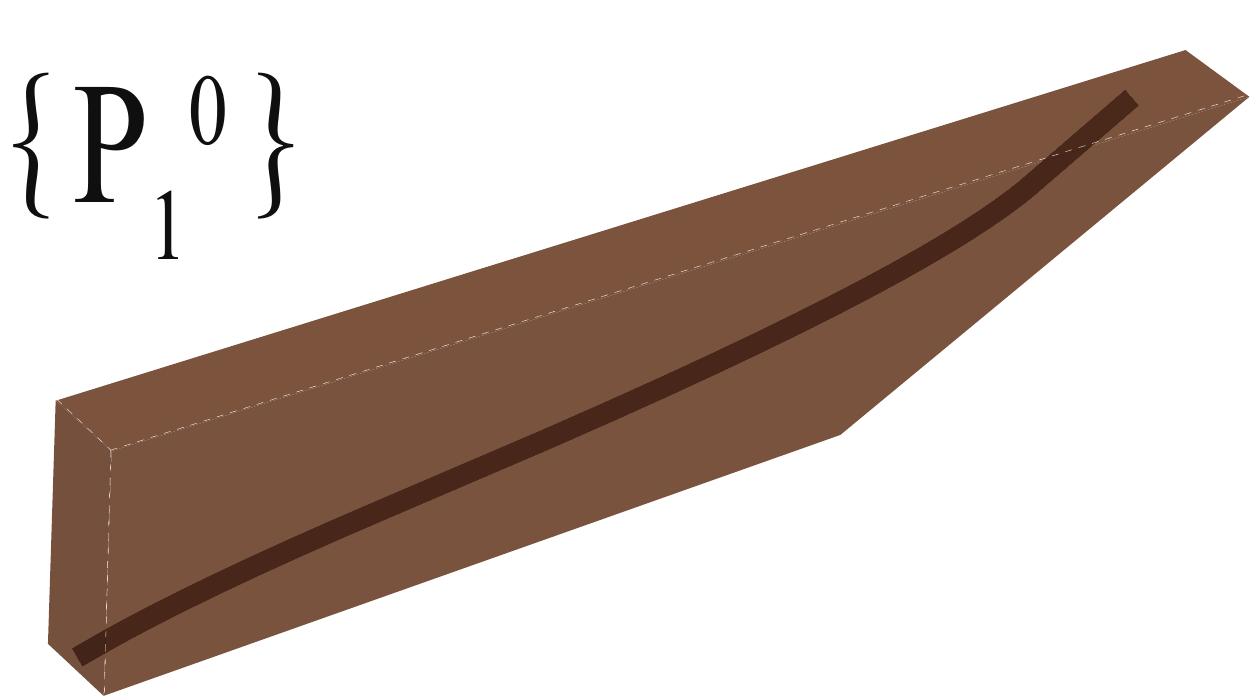}
	\end{minipage}}
	\subfigure[\myadd{To left.}]{	\begin{minipage}[b]{0.3\linewidth}
			\includegraphics[width=1.\columnwidth]{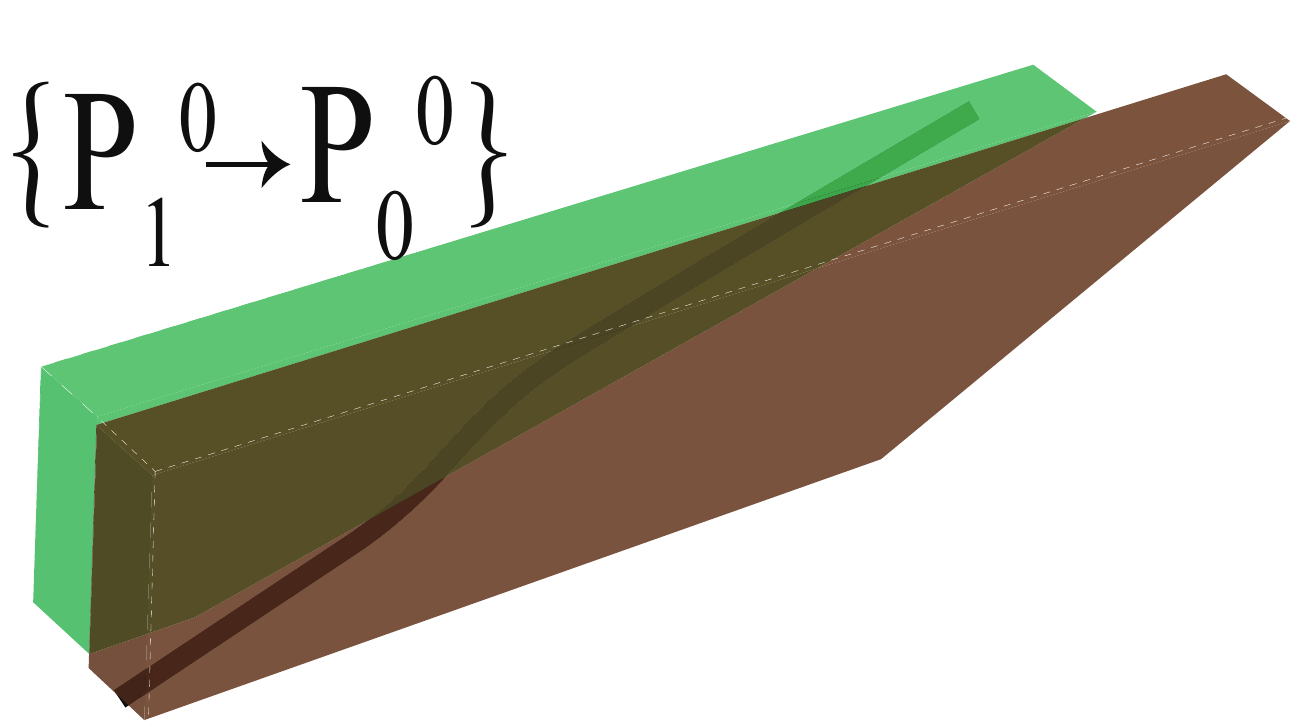}
	\end{minipage}}
	\subfigure[\myadd{To left.}]{	\begin{minipage}[b]{0.3\linewidth}
			\includegraphics[width=1\columnwidth]{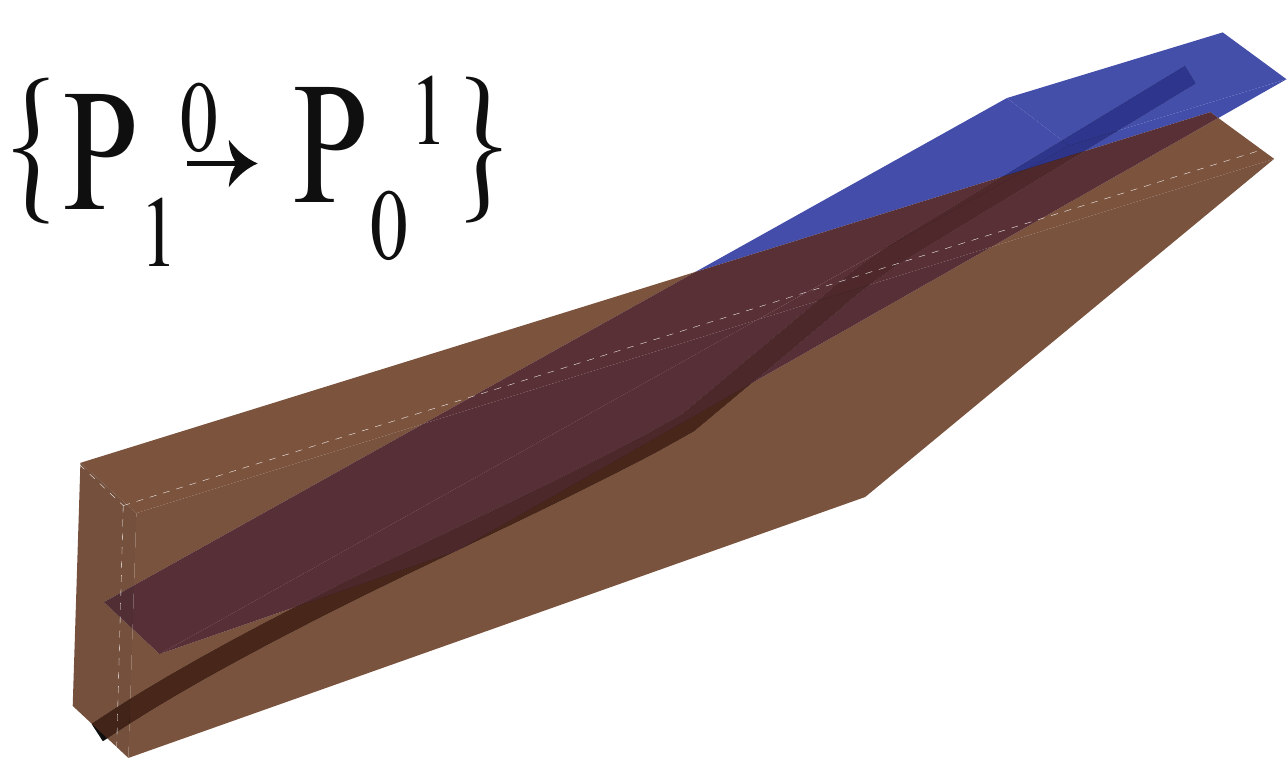}
	\end{minipage}}
	\subfigure[\myadd{Back after to left.}]{	\begin{minipage}[b]{0.3\linewidth}
			\includegraphics[width=1\columnwidth]{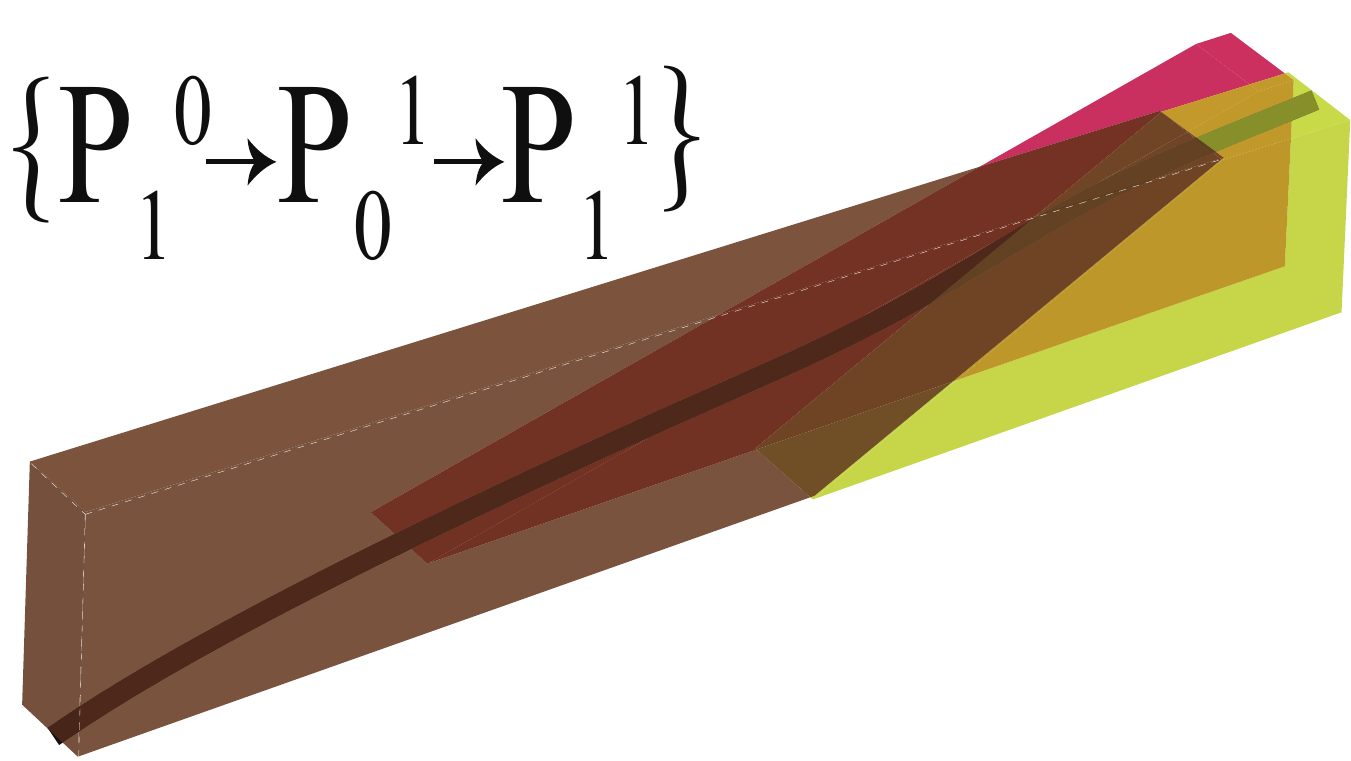}
	\end{minipage}}
	\subfigure[\myadd{To left.}]{	\begin{minipage}[b]{0.3\linewidth}
			\includegraphics[width=1.\columnwidth]{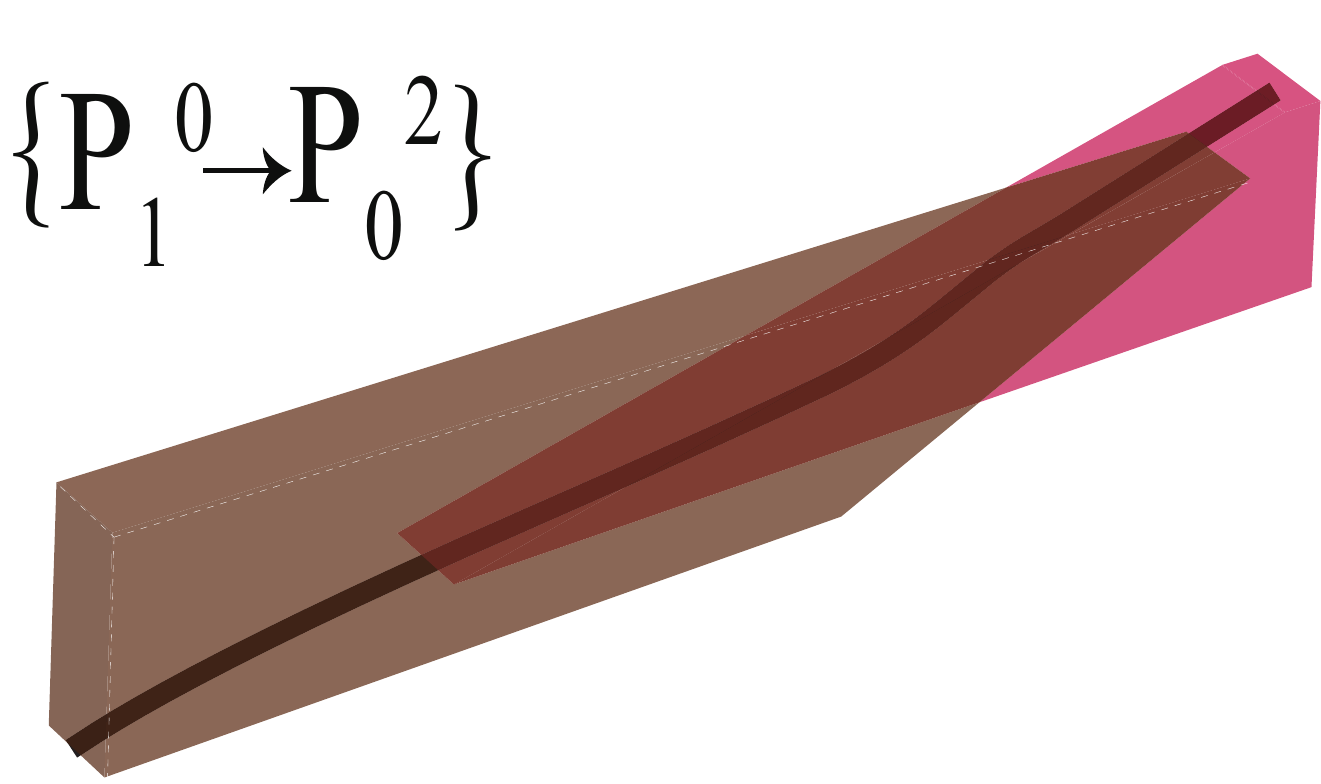}
	\end{minipage}}
	\subfigure[\myadd{Back after to left.}]{	\begin{minipage}[b]{0.3\linewidth}
			\includegraphics[width=1.\columnwidth]{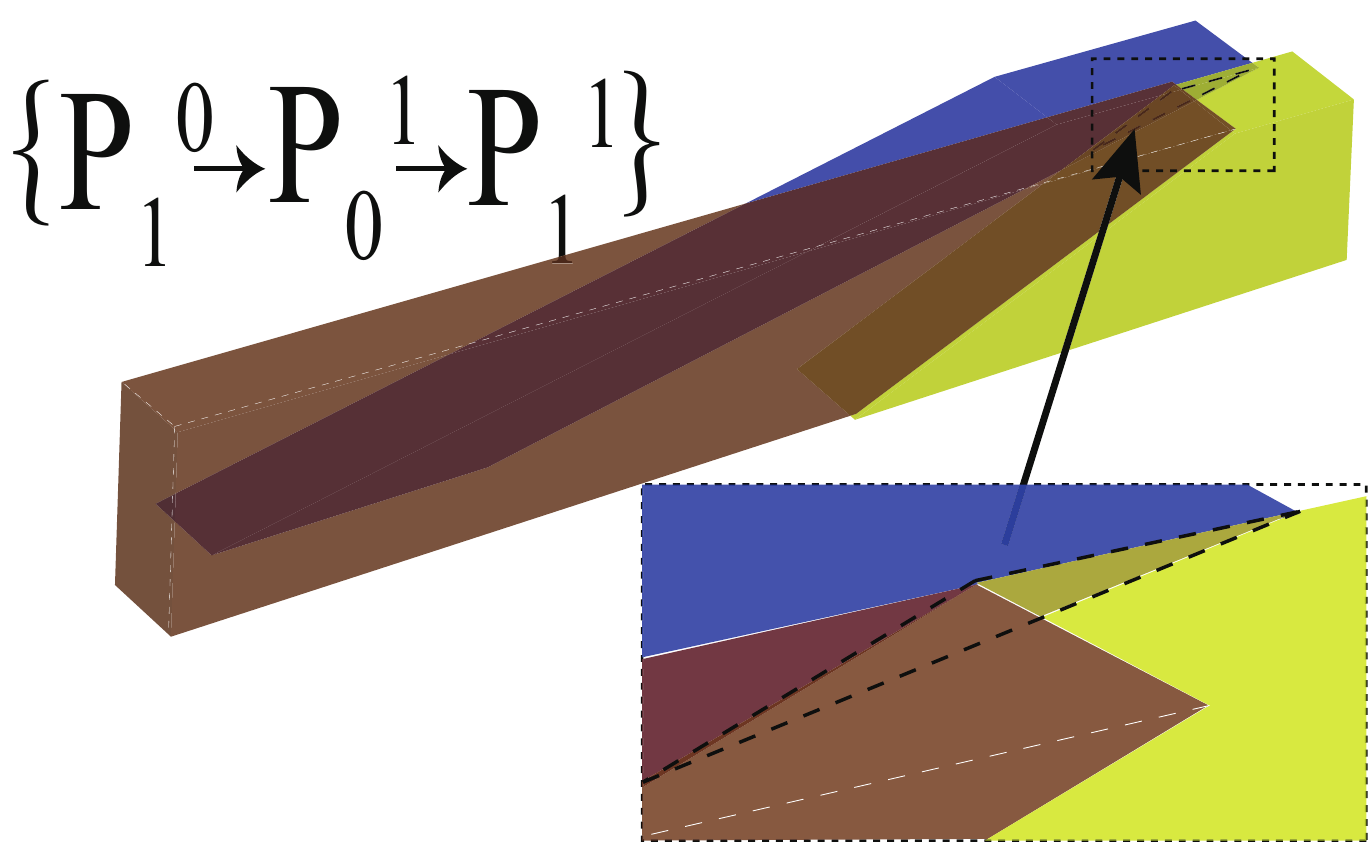}
			\label{homotopy:maneuver6}
	\end{minipage}}
	
	\caption{\myadd{Homotopy classes consider temporal dimension. Profiles are created by the vehicles in Fig.}\ref{homotopy:dynamic-vehicles}. \myadd{In Fig.}\ref{homotopy:maneuver6}\myadd{, the time interval between profile $ P_{0}^{1} $ and profile $ P_{1}^{1} $ is too narrow to change to the right lane.}}
	\label{homotopy:dynamic-homotopy}
\end{figure}

Considering the temporal aspect of vehicles, a different kind of homotopy paths exists in the spatio-temporal without collision with other vehicles. In Fig.\ref{homotopy:dynamic-vehicles} two vehicles in the left lane yield three alternative collision-free spaces and the yellow vehicle splits right lane into two halves. Primitively, connecting the center of a vehicle at each time instance  by a single line generates the future trajectory $ \mathbf{x} $. A plane is expanded by $ \mathbf{x} $ and\myremove{a line perpendicular to both $ \mathbf{s} $ coordinate and $ \mathbf{x} $}\myadd{$ d $-axis.}The plane splits spatio-temporal space into two halves. Each space is defined as a Trajectory Profile (TP), that is, a profile $ P_{i}^{j} $ is the $ j^{th} $ collision-free zone split by different vehicles without a hole in the $ i^{th} $ lane, where $ i $ and $ j $ $\in$ \{0, 1, 2, 3, \ldots \}. Fig.\ref{homotopy:dynamic-homotopy} reappears the 3D view of TPs in Fig.\ref{homotopy:dynamic-vehicles}. The green profile $ P_{0}^{0} $ is the space just behind the blue vehicle in the left lane, and the yellow profile $ P_{1}^{1} $ is the space in front of the yellow vehicle. The purple trajectory taking over the red and the yellow vehicles successively arrives at the front region of the yellow vehicle. The corresponding maneuver can be represented by a sequence of connective profiles $ \{ P_{1}^{0} \rightarrow P_{0}^{2} \rightarrow P_{1}^{1} \} $, which means a unique homotopy class.

\section{Topology Generation}
\label{sec_topology_generation}
To enable an enumeration of maneuvers, a topological structure of the agent's planning space must be generated. There are three basic steps needed. Firstly, we propose an algorithm of generating topological corridors by considering obstacles that limit the non-collision spatial space of the agent in the lanes. Then an algorithm of generating topological routes is presented using the mobility of our agent and future trajectories of other vehicles to split the whole spatio-temporal space. Finally, by matching the corridors and routes in lane level, we enumerate all possible maneuvers in the predicting horizon.

\subsection{Topology by Static Obstacles}
\label{Topological_static}
As described in \ref{homotopy-decsribtion}, the agent has more choices to pass the occupied region when the obstacles block part of the lane. However, fewer options exist when the whole right lane is occupied in a double-lane road. The disordered and irregular obstacles give rise to the complexity of planning space. Firstly, the whole space are split into pieces, then, they are reconnected into several structured spaces where trajectory planning is easier.

\subsubsection{Lane Split and Reconnection}
For being able to decide discretely on which side of a lane that the agent can go through, we split the lane into seven parallel bands based on experience. And the width of each band is $ W_{b}= W_{lane}/7 $, where $ W_{lane} $ is the lane\textquotesingle s width. Each band is bilaterally shifted from the\myremove{lane}center \myadd{of lane.}The center of middle band is the center of lane.\myremove{And}The minimum width of drivable side $ W_{d} $, is
\begin{equation}
W_{d} = \max( 3W_{b} , W_{vehicle}+ 2d_{safe})\label{eq:band_width}
\end{equation}
where $ d_{safe} $ is the safe distance to obstacles, and $ W_{vehicle} $ is the width of a vehicle.
\begin{figure}[htbp]
	\centering
	\subfigure[Corridor 1.]	{
		\begin{minipage}[t]{0.22\linewidth}
			\includegraphics[width=1\columnwidth]{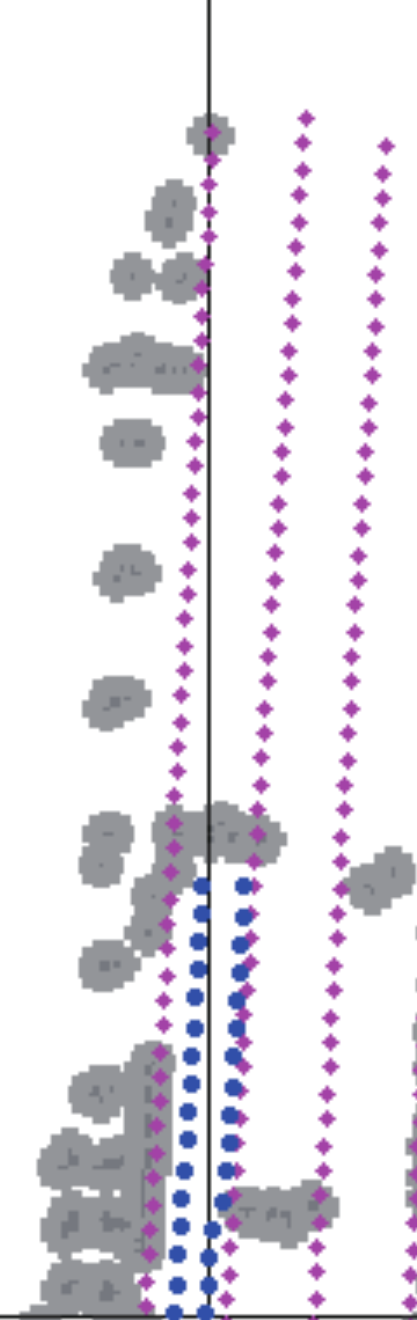}
	\end{minipage}
	}
	\subfigure[Corridor 2.]{
		\begin{minipage}[t]{0.22\linewidth}
			\includegraphics[width=1\columnwidth]{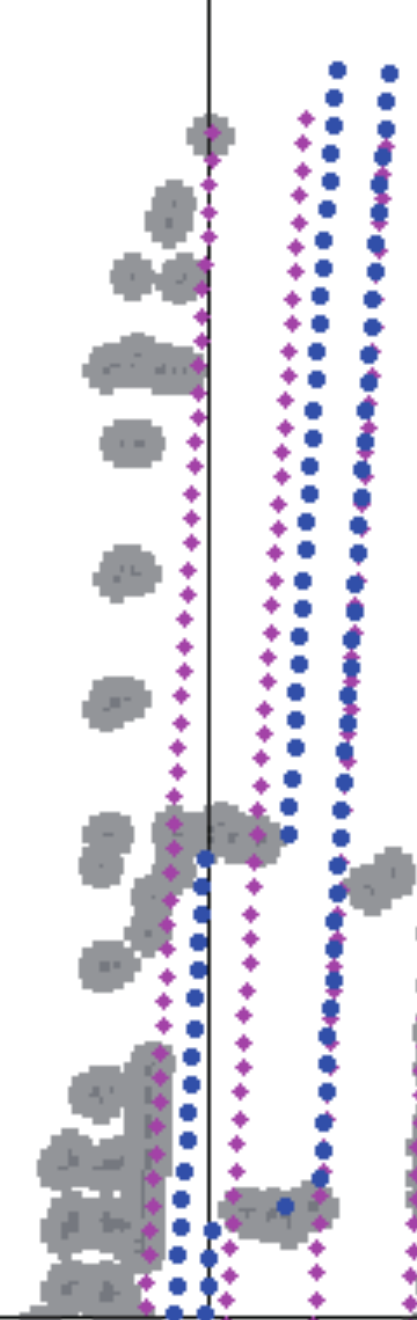}
	\end{minipage}}
	\subfigure[{Corridor 3.}]{
	\begin{minipage}[t]{0.22\linewidth}
		\includegraphics[width=1\columnwidth]{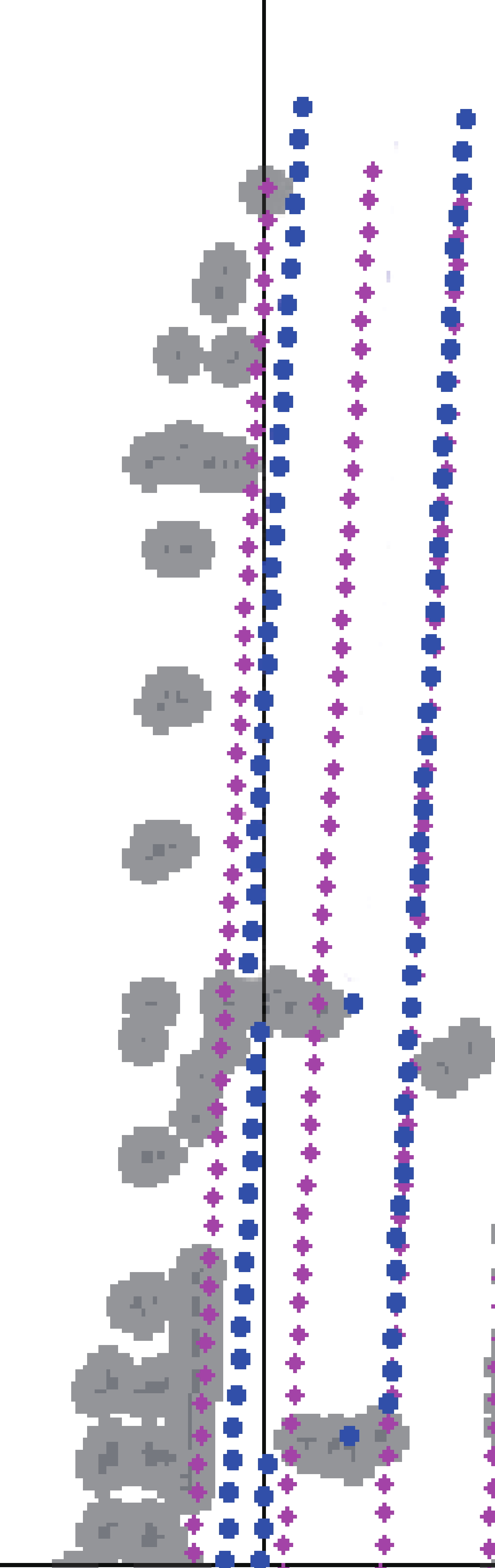}
	\end{minipage}}
	\subfigure[Band split.]{
		\begin{minipage}[t]{0.22\linewidth}
			\includegraphics[width=1\columnwidth]{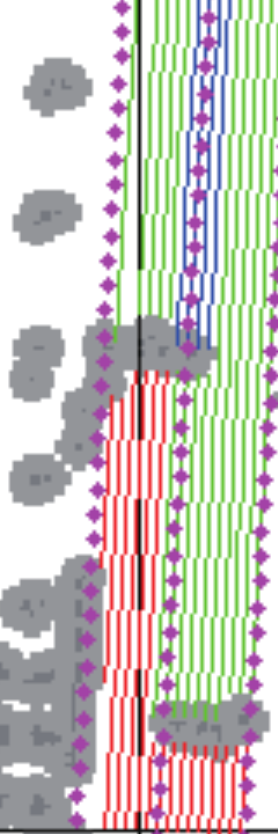}
	\end{minipage}}
	\label{sections:timeA}
%
	\caption{Homotopy paths in a double-lane road with obstacles. The red dots represent lane lines, and the blue dots represent left and right boundaries of a corridor. The colors of three sections in one band are red, cyan and blue respectively. \myadd{The dark grey dots are the obstacles, and the light grey areas are the inflated area of obstacles.}}
	\label{sections:all}
\end{figure}

\begin{figure}[!htbp]
	\subfigure[Corridors starts in the left lane.]{
	\begin{minipage}[t]{0.48\linewidth}
		\includegraphics[width=1\columnwidth]{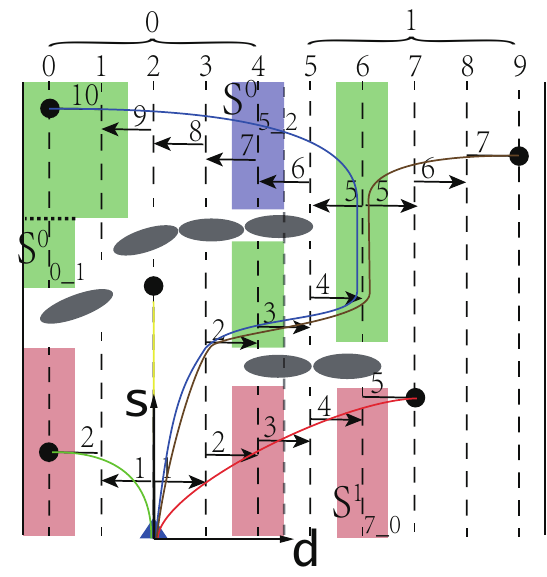}	
		\label{EndSections:left}
	\end{minipage}}
	\subfigure[Corridors starts in the right lane.]{
	\begin{minipage}[t]{0.48\linewidth}
		\includegraphics[width=1\columnwidth]{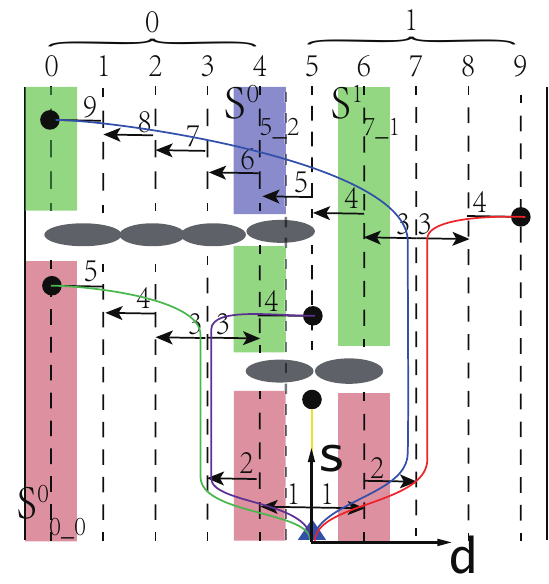}
		\label{EndSections:right}
	\end{minipage}}
	\caption{Terminal sections generated in two similar scenarios with different initial positions in two lanes. The start pose locates at the blue triangle. The grey areas are the inflated obstacles. The light red, cyan and blue bands are the corresponding sections in Fig.\ref{sections:all}. The arrow line with a number means the search direction with depth. The black dot at the end of each line represents a terminal section.}
	\label{EndSections:all}
\end{figure}

As shown in Fig.\ref{sections:all}, the obstacles occupy the lane permanently and cut one band into several sections. One section $ s_{j\_k}^{i} $ is the $ k_{th} $ section in $ j_{th} $ band, and $ i $ represents lane id. Where i $ \in(0, 1 \dots N), j \in(0, 1, \dots ,7N) $ and $ k \in(0, 1, 2) $.\myremove{Here at}\myadd{At}most three sections in each band of N lanes are considered, since the smoothness of a lane is directly related to the distribution of the obstacles. An illustrative example is also shown in Fig.\ref{EndSections:all}, and the lane is split  into five bands for the simplicity of explaining. Only the front two sections are enough to plan a reasonable path according to experience. The advantages are two folds. Firstly, the split is only related to the obstacles but not a designed sampling distance. The split implies a more natural choice of the sampling interval, thus avoiding an unevenness path from fixed  value of sampling distance. Secondly, the number or the length of sections in one line directly reflects the smoothness of the front road, and this will guide the agent at what level the speed should be.
 
One generated section is viewed as a node while searching. One section connects only to those in adjacent bands. Given no more than three sections in each band, there are at most five edges connecting two bands. Sections are connected in a directed graph following rules \myadd{below}
\newcounter{Lcount}
\begin{list}{\textbf{ Rule {\arabic{Lcount}}}:}	{\usecounter{Lcount} \leftmargin=0em}
	\item The connection means that the adjacent sections have a minimum overlapping thresh (the width of the agent) along lane direction.\label{rule_connection_min}
	\item Connections start from \textit{root section} (where vehicle locates at) to the leftest and rightest sides separately.\label{rue_search_direction}
	\item Different sections in the same line cannot connect to each other directly. The connection to different sections in one band has at least one transitional section in a adjacent line.\label{rule_connect_adjacent}
\end{list}
A directed graph of sections is generated once the connection work is done. The procedure is shown in Fig.\ref{EndSections:all}.

\algnewcommand\algorithmicforeach{\textbf{for each}}
\algdef{S}[FOR]{ForEach}[1]{\algorithmicforeach\ #1\ \algorithmicdo}
\algnewcommand\algorithmicsuchthat{\textbf{such that}}
\algdef{S}[FOR]{ForSuch}[2]{\algorithmicforeach\ #1\ \algorithmicsuchthat\ #2 \algorithmicdo }
\begin{algorithm}
	\caption{Generate Topological Corridors in 2D Space}\label{euclid}
	\label{alg:2D_algorithm}
{\small{	\begin{algorithmic}[1]
		\Require cost-map:$ m_{cost} $, lane lines:$ L_{lines} $
		\Ensure corridors with different homotopy type:$ C_{cor} $

		\State $ G_{sec}$ $\gets$ \textit{generateSectionGraph($ m_{cost}, L_{lines} $)}
		\label{alg:2D:core-gensection}
		\State $ C_{sec} $ $\gets$ \Call{generate-end-sections}{$ G_{sec} $, $ s_{r} $}
		\label{alg:2D:core-endsections}
		\State $ C_{cor} $ $\gets$ \Call{generate-Corridors}{$ C_{sec}, G_{sec} $}
		\label{alg:2D:core-genCorridors}
		
		\Statex
		\Function {generate-end-Sections}{$ G_{sec} $, $ s_{r} $}
			\label{alg:2D:section-ends-generation}
			\State {${Init} $: \textit{unvisited} sections $  G_{uv}\gets G_{sec} $,  $ C_{sec} \gets s_{r} $ }
			\While { $ G_{uv} \neq \emptyset $ }
				\ForEach { $ s_{c} \in C_{sec} $} \Comment{$ s_{c} $ is \textit{current section}}
					\State $ s_{n} \gets s_{n(i)} \cap s_{c} \neq \emptyset$, $ i \subset[0,4] $
					\State $ G_{uv} \gets G_{uv}\backslash s_{c} $ 
					\If{ $ s_{n} \neq \emptyset$ }
						\State $ G_{sec} \gets G_{sec}\backslash s_{c} $
						\State $ G_{sec} \gets G_{sec}\cup \{s_{n}\} $
					\EndIf
				\EndFor		
			\EndWhile
			\If { $ s_{r} \notin G_{sec} $}
				\State $ G_{sec} \gets G_{sec}\cup\{s_{r}\}  $	
			\EndIf
			\State \textbf{return} $ C_{sec} $
		\EndFunction
		\Statex
		\Function {generate-Corridors}{$ C_{sec}, G_{sec} $}
			\State ${Init}: $ \textit{Corridors}:$ C_{cor} \gets \emptyset $
			\ForEach { $ s_{c} \in C_{sec} $}
			\label{alg:2D:remove-unreasonable-leaf}
				\If {$ s_{c} $ is much shorter than \textit{parent section}  }
					\State $ s_{c}\gets s_{p} $  \Comment{move to \textit{parent section}} 
				\EndIf
			\EndFor
			\ForEach {  $ s_{c} \in  C_{sec} $}
			\label{alg:2D:generate-corridor-inversely}
				\State$ c_{cor} $ $\gets$ \textit{connect} from $ s_{c} $ to $ s_{r} $ \Comment{get a corridor}
				\State \textit{add} $ c_{cor} $ to $ C_{cor} $
			\EndFor
				
			\ForEach { $ c_{cor} \in C_{cor} $ }
			\label{alg:2D:complement-and-smooth}
				\ForEach{ $ \bar{c}_{cor} \in C_{cor}{ }\ \&\ c_{cor} \neq \bar{c}_{cor}$}
					\State \textit{complement} $ c_{cor} $ with $ c_{cor-next} $
				\EndFor
				\State $ c_{cor} \gets $ \textit{smoothSectionInCorridor($ c_{cor} $)}
				\label{alg:2D:smooth}
			\EndFor 
			\State \textbf{return} $ C_{cor} $
		\EndFunction
	\end{algorithmic}
}}
\end{algorithm}

\subsubsection{Generate topological corridors in 2D space}
\textbf{Algorithm~\ref{alg:2D_algorithm}} shows high-level pseudo-code for the generation of topological corridors at each planning\myremove{circle}\myadd{cycle.}The directed graph of sections, $ G_{sec} $, is created in line \text{\ref{alg:2D:core-gensection}} given lane lines as bounds and a cost-map generated from obstacles. As shown in Fig.\ref{sections:all} and \ref{EndSections:all}, the connection between sections of adjacent lines is checked according to \textit{Rule \ref{rule_connection_min}}.

One core function in line \text{\ref{alg:2D:core-endsections}} finds the terminal sections of all possible topological corridors $ C_{sec} $. Originally, the $ C_{sec} $ is only initialized with \textit{root section} $ s_{r} $ (the blue triangle). The corridor searching starts from each element in $ C_{sec} $ in every loop. If the current section $ s_{c} $ in $ C_{sec} $ connects to adjacent sections, then the connected sections are added into $ C_{sec} $ for next loop quire and current section $ s_{c} $ is removed (see line \ref{alg:2D:section-ends-generation}). Looping stops until all sections have been visited, and the elements left in $ C_{sec} $ represent all potential topological corridors. Taking Fig.\ref{EndSections:left}, for example, searching starts with only \textit{root section} $ s_{2\_0}^{0} $ in $ C_{sec} $. The next step confirms two connections that $ s_{1\_0}^{0} $ and $ s_{3\_0}^{0} $ are connected to $ s_{2\_0}^{0} $, then $ C_{sec} $ is updated with $ s_{1\_0}^{0} $ and $ s_{3\_0}^{0} $ instead of $ s_{2\_0}^{0} $. In the next loop, $ s_{0\_0}^{0} $ taking place of $ s_{1\_0}^{0} $ becomes a real terminal section. Besides, unvisited sections $ s_{4\_0}^{0} $ and $ s_{4\_1}^{0} $ connect to $ s_{3\_0}^{0} $ respectively and take place of $ s_{3\_0}^{0} $ in $ C_{sec} $. The search ends with four real terminal sections $ s_{0\_0}^{0} $, $ s_{7\_0}^{1} $, $ s_{9\_0}^{1} $, $ s_{0\_1}^{0} $ and one extra \textit{root section} $ s_{2\_0}^{0} $. 

Procedure above mainly finds a possible path to the terminal section without left and right boundary constraints. The other core function in line \text{\ref{alg:2D:core-genCorridors}} generates real drivable corridors. 
Due to the irregular shape and the complex distribution of the obstacles, some unreasonable corridors are inevitable. Illogical terminal sections are removed inside the first loop in line \ref{alg:2D:remove-unreasonable-leaf}. The next loop body searches a corridor from terminal section to \textit{root section} $ s_{r} $ reversely. One problem still exists as the sections stretch from current position to either side, thus the corridor is unable to completely coverage a total lane. A quick solution is to query from sections of other corridors that connecting to either side of the current corridor. For this reason, corridors with missing part can be easily complemented. As illustrated in Fig.\ref{EndSections:left}, an incomplete corridor in red curve covers six sections in bands from 2 to 7. This corridor leads vehicle to Lane \textbf{1} before passing the obstacles and the uncovered sections of lane 1 can be complemented by querying from the corridor in brown curve that ends at $ s_{9\_0}^{1} $. For the convenience to decide the width of corridor, some protuberant bands are truncated to fit adjacent bands. For example, the terminal section $ s_{0\_1}^{0} $ in blue curve is trimmed off to match with parent section $ s_{1\_1}^{0} $.  Now all the drivable corridors are generated without considering the vehicles.

The corridor generation algorithm has three basic advantages as below. Firstly, the number and the length of sections are qualitative descriptions of the road smoothness. Secondly, the agent\textquotesingle s position and obstacle distribution always implicitly ignore those corridors crossing more lateral bands. In Fig.\ref{EndSections:right}, a hidden corridor ($ s_{5\_0}^{1} $$\rightarrow$$ s_{3\_0}^{0} $$\rightarrow$$ s_{6\_1}^{1} $$\rightarrow$$ s_{0\_1}^{0} $) is ignored naturally, as the blue route reaches section $ s_{6\_1}^{1} $ at third step firstly.  And the hidden corridor is taken over since it takes\myremove{more}\myadd{a longer}lateral shift distance to arrive at the same position. Finally, the corridors are generated with only one time collision checking in the $ s $\nobreakdash--$ d $ plane at the phase of lane split. By simply querying from the sections of other corridors, the lateral coverage of a corridor is complemented easily. Hence, the algorithm generates corridors in different homotopy classes quickly with all lane-level terminal regions reachable. 

\subsection{Topology by Dynamic Vehicles}
\label{Topological_dynamic}
The moving vehicles make both maneuver searching and trajectory planning problems more complicated. Different from the process of the stationary obstacles in $d$\nobreakdash--$s$ plane, we parse the predicted trajectories of dynamic vehicles in $ s $\nobreakdash--$ t $ plane. 

One reasonable agent should seek the driving ability of human beings, which will be more acceptable.
Human drivers pay special attention to the near front vehicles in surrounding lanes to avoid collision, and people tend to keep a safe distance to front vehicle. Accordingly, the generating algorithm of topological routes are proposed based on the following assumptions: 
\begin{list}{ \textbf{Assumption {\arabic{Lcount}}}:} {\usecounter{Lcount}\leftmargin=0em}
	\item A front accelerating vehicle is treated as a constant speed one in the predicting time, but others vehicle are treated faithfully. \myadd{If the acceleration information or the future position states of the front vehicle can be acquired accurately, the vehicle can be treated faithfully too.}\label{assum_trajectory}
	\item The nearest front vehicle in each lane is viewed as a \textit{reference vehicle}. If the front vehicle does not exist, the nearest back one in the same lane is a \textit{reference vehicle}.\label{assum_reference_vehicle}
	\item A vehicle always occupies the whole lane it is in, so one vehicle must take over other vehicles in adjacent lanes. A vehicle changing lane occupies two lanes at one time.\label{constraint_occupy_lanes}
\end{list}

\begin{figure}[htbp]
	\centering
	\includegraphics[width=0.9\columnwidth]{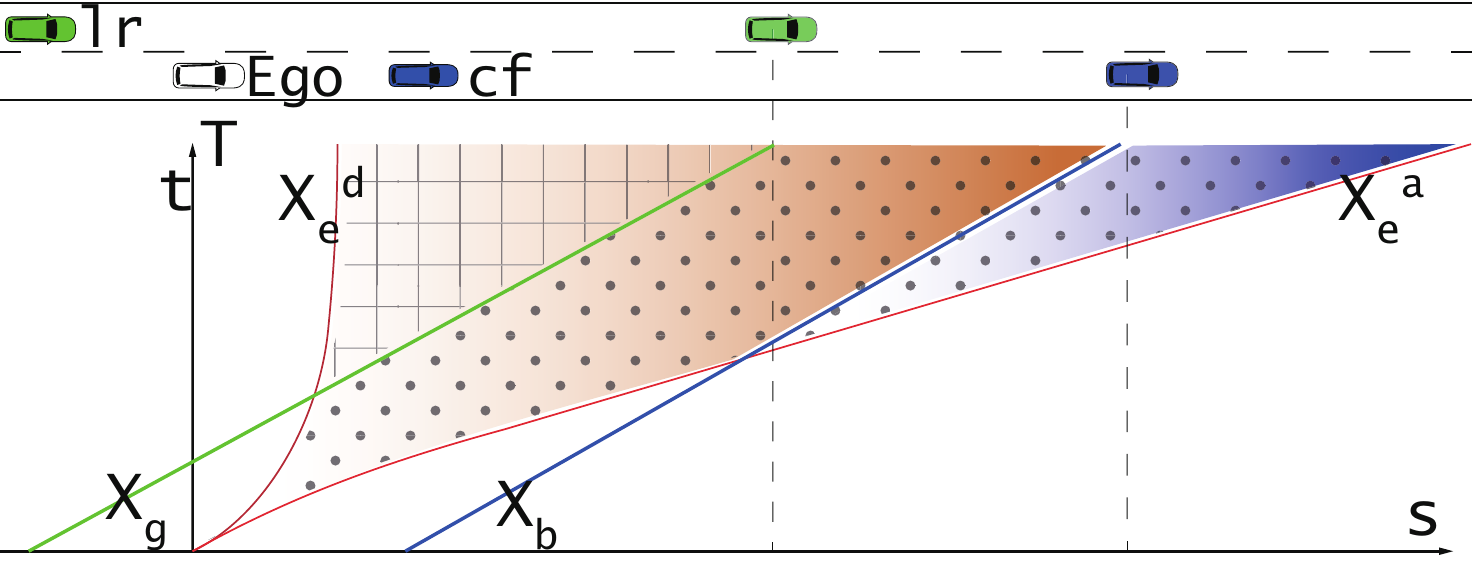}
	\caption{\myadd{TPs in two lane situation considering vehicular mobility in side view. The mesh and dot filled profiles are split by $ \mathbf{x}_{g} $ (trajectory of the green vehicle). The brown and blue profiles are split by $ \mathbf{x}_{b} $ (trajectory of the blue vehicle). $ \mathbf{lr} $ denotes the left rear vehicle in green.}}
	\label{Dynamic:basic}
\end{figure}


\subsubsection{Constraints of vehicular mobility}
An agent has its limitations of both deceleration and acceleration at different velocities. The values can be quite different for  vehicle to vehicle. However for a specific vehicle or driving strategy, deceleration and acceleration can be mapped to velocity by functions $ f_{d}(v) $ and $ f_{a}(v) $ without considering load fluctuation and road condition. As show in Fig.\ref{Dynamic:basic}, the red curve $ \mathbf{x}_{e}^{d} $ represents the trajectory of the agent decelerating to zero velocity. $ \mathbf{x}_{e}^{a} $ represents the trajectory of accelerating the agent to the permitted maximum velocity. Obviously, the planning space should be limited between $ \mathbf{x}_{e}^{d} $ and $ \mathbf{x}_{e}^{a} $. The TP limited by $ \mathbf{x}_{e}^{d} $ and $ \mathbf{x}_{e}^{a} $ are defined as the \textit{base profile} $ P_{b} $. $ \mathbf{x}_{r} $ and $ \mathbf{x}_{f} $ represents the rear and front boundaries of a TP respectively.

\subsubsection{Spatio-temporal space split}
\begin{figure}[htbp]
	\centering
	\includegraphics[width=0.9\columnwidth]{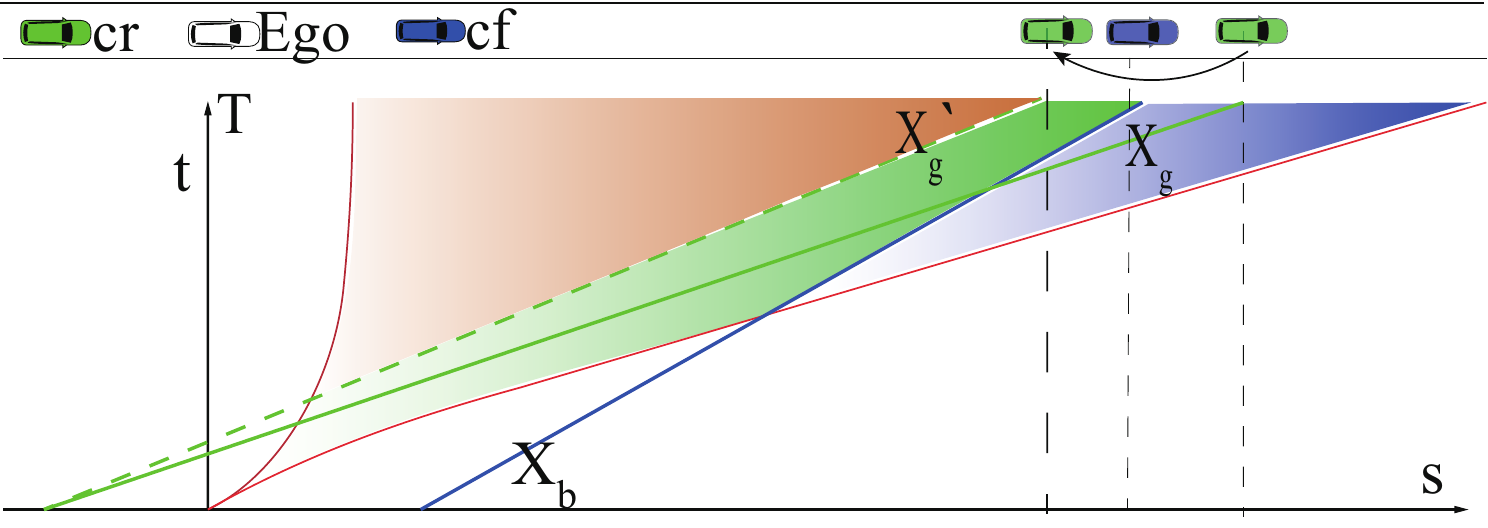}
	\caption{\myadd{Example of adjusting the trajectory of a \textit{non-reference vehicle} in side view.}}
	\label{Dynamic:adjust}
\end{figure}

Before splitting the spatio-temporal space, whether surrounding vehicles will intersect with the lanes that the agent is in and will be in are checked. Vehicles that will not affect\myremove{our}\myadd{the}agent in predicting time are ignored, and it means no any part of their trajectories locate between $ \mathbf{x}_{e}^{d} $ and $ \mathbf{x}_{e}^{a} $. All relevant vehicles are permuted from back to front in each lane according to assumption of a \textit{reference vehicle}. Taking Fig.\ref{Dynamic:basic} for an illustrative example, $ V_{lr} $ and $ V_{cf} $ are both \textit{reference vehicles}, as $ V_{cf} $ is the nearest front vehicle in current lane and $ V_{lr} $ is the only vehicle in the left lane. But $ V_{cr} $ in Fig.\ref{Dynamic:adjust} is not a \textit{reference vehicle}, since $ V_{cf} $ is in front of the agent in the same lane. 

The agent should maneuver according to the \textit{reference vehicle} $ V_{cf} $ in priority. So the trajectories of \textit{non-reference vehicles}  are adjusted to guarantee no intersections between trajectories of other traffic participants. Just as $ V_{cr} $ in Fig.\ref{Dynamic:adjust}, originally $ V_{cr} $ will pass through $ V_{cf} $ in trajectory $ X_{g} $, but this is illogical. 
$ V_{cf} $ is moved to the back of blue vehicle in trajectory $ X_{g}^{'} $ and no intersection exists between trajectories of two vehicles anymore. To maintain the situation of traffic, the moved distance is strictly limited.\myremove{And}Now the trajectory should be generated in the relatively narrow TP of\myremove{gradient}green \myadd{gradient.}

\subsubsection{Trajectory Profiles generation and connection}
We use a route $ r_{rt} = [s_{0},\dots, s_{N}] \in \mathbf{R}_{rt}$  to represents a path with lateral pose uncertain in $ s $\nobreakdash--$ t $ plane, where N is the maximum time index for the route. A route $ r_{rt} $ can be represented by one TP (where vehicle locates in) or a sequence of connective TPs. Algorithm \ref{alg:3D_algorithm} performs the main procedure for the generation and connection of TPs.

\renewcommand{\algorithmicrequire}{\textbf{Input:}} 
\renewcommand{\algorithmicensure}{\textbf{Output:}} 
\begin{algorithm}[!ht]
	\caption{Generate Topological Routes in 2D Space}
	\label{alg:3D_algorithm}
	{\small{	\begin{algorithmic}[1]
			\Require permuted trajectories $ \mathbf{x}_{k}\subset\mathbf{X}, k\in \{0, 1, 2,\ldots \}$ and \textit{base profile} $ P_{b} $
			\Ensure routes with different homotopy type: $ \mathbf{R}_{rt} $
			
			\State $ \mathbf{P}_{pfl} $ $\gets$  \Call{generate-Profiles}{$ \mathbf{X} $}
			\label{alg:3D:core-generate-profiles}
			\State $ \mathbf{R}_{rt} \gets $ \Call{connect-Profiles}{$ \mathbf{P}_{pfl} $}
			\label{alg:3D:core-connect-profiles}
			
			\Statex
			\Function {generate-Profiles}{$ X_{i} $}
			\label{alg:3D:generate-profiles}
			\State {${Init}:\ $profiles \textit{set} $ \mathbf{P}_{pfl} \gets \emptyset $, \textit{parent profile} $ {P}_{p} \gets {P}_{b}$ }
			\ForEach { $ l_{i} $, $ i\in \{0, 1, 2\}$} \Comment{$ l_{1} $:current lane}
			\ForEach {trajectory $ \mathbf{x}_{k} $}
			\State $\{{P}_{i}^{j}\}  \gets$ \textit{split} $ {P}_{p} $ with $ \mathbf{x}_{k} $, $ j \in [0,2] $
			\If {$ P_{i}^{j} $ is \textit{narrow} { }\&{ } the agent \textit{not in} $ P_{i}^{j} $}
			\State $ \{{P}_{i}^{j}\} \gets \{{P}_{i}^{j}\}\backslash P_{i}^{j} $ \label{alg:3D:eliminate_profile}
			\EndIf
			\State $ {P}_{p} \gets P_{i}^{j}\mid $\textit{below} $ \mathbf{x}_{k} $, with max arrival distance
			\State $ \mathbf{P}_{pfl} \gets \mathbf{P}_{pfl} \cup\{{P}_{i}^{j}\} $
			\EndFor
			\State $ \mathbf{P}_{pfl} \gets \mathbf{P}_{pfl} \cup\{{P}_{p}\} $
			\EndFor \\
			\Return $ \mathbf{P}_{pfl} $
			\EndFunction
			
			\Statex
			\Function {connect-Profiles}{$ \mathbf{P}_{pfl} $}
			\label{alg:3D:connect-profiles}
			\State {${Init}: $ routes $ \mathbf{R}_{rt}\gets P_{ego}$ } \Comment{$  P_{ego} $ is \textit{root profile}}
			\For {$ i =1 $ to $ n_{depth} $} 	\Comment{$ n_{depth } = 3 $}
			\For {each $ r_{rt} $ in $ \mathbf{R}_{rt} $}
			\State $ P_{m}^{n} \gets $ \textit{end profile} of $ r_{rt} $
			\ForEach {$ i\in\{m-1, m+1\} $}
			\State $ P_{i}^{j} \gets P_{i}^{j}\cap P_{m}^{n} \neq \emptyset ${ }\&{ }$ P_{i}^{j} $ \textit{not in} $ r_{rt} $ \label{alg:3D:connect-overlapping}
			\State $ \bar{r}_{rt} \gets $ \textit{append} $ P_{i}^{j} $ to $ r_{rt} $
			\State $ \mathbf{R}_{rt}\gets \mathbf{R}_{rt}\cup \{\bar{r}_{rt}\}$ 
			\EndFor
			\EndFor
			\EndFor \\
			\Return $ \mathbf{R}_{rt} $
			\EndFunction
		\end{algorithmic}
	}}
\end{algorithm}

\begin{figure}[htbp]
	\centering
	\subfigure[One crunode at $ \mathbf{x}_{f} $ ]{	\begin{minipage}[t]{0.31\linewidth}
		\includegraphics[width=1\columnwidth]{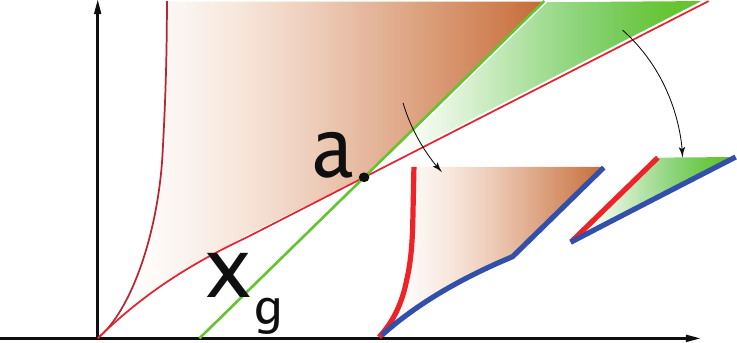}		
		\label{profiles_generation_a}
	\end{minipage}}
	\subfigure[One crunode at $ \mathbf{x}_{r} $]{	\begin{minipage}[t]{0.30\linewidth}
		\includegraphics[width=1\columnwidth]{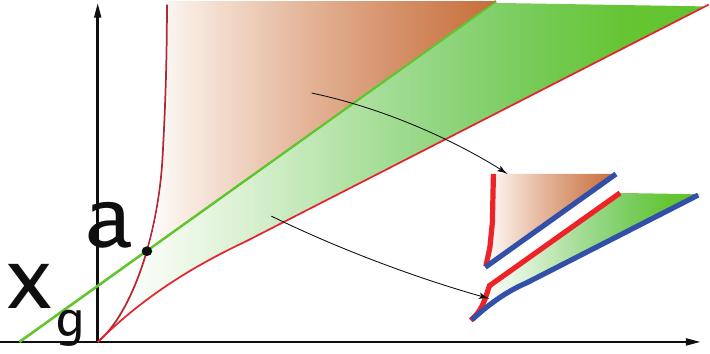}		
		\label{profiles_generation_b}
	\end{minipage}}
	\subfigure[Crunodes at both \textbf{1}]{	\begin{minipage}[t]{0.30\linewidth}
		\includegraphics[width=1\columnwidth]{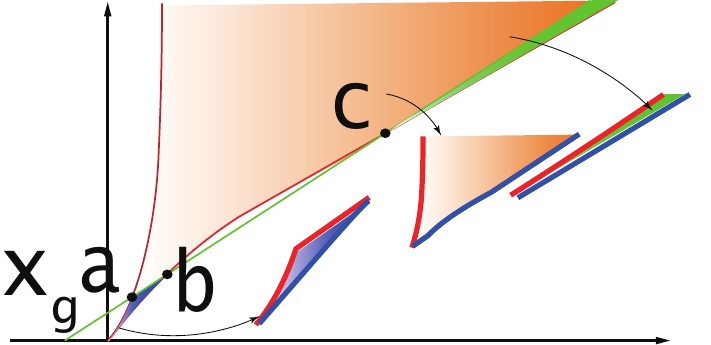}		
		\label{profiles_generation_c}
	\end{minipage}}

	\subfigure[Crunodes at both \textbf{2}]{	\begin{minipage}[t]{0.31\linewidth}
		\includegraphics[width=1\columnwidth]{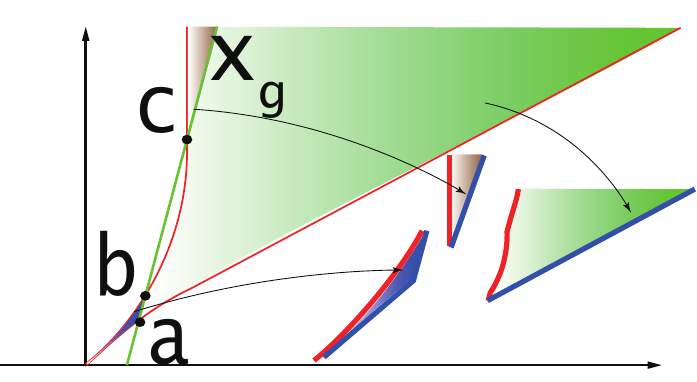}		
		\label{profiles_generation_d}
	\end{minipage}}
	\subfigure[Crunodes at both \textbf{3}]{	\begin{minipage}[t]{0.32\linewidth}
		\includegraphics[width=1\columnwidth]{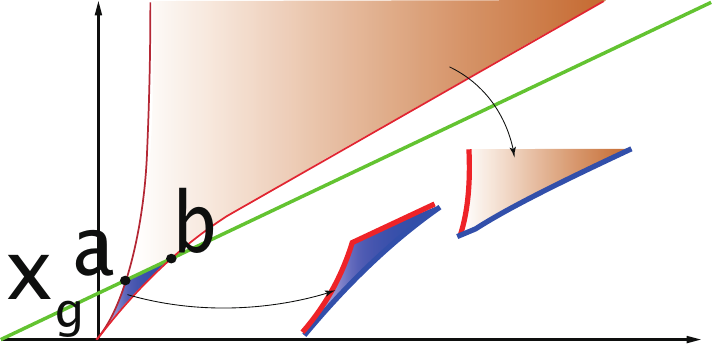}		
		\label{profiles_generation_e}
	\end{minipage}}
	\subfigure[Two trajectories]{	\begin{minipage}[t]{0.31\linewidth}
		\includegraphics[width=1\columnwidth]{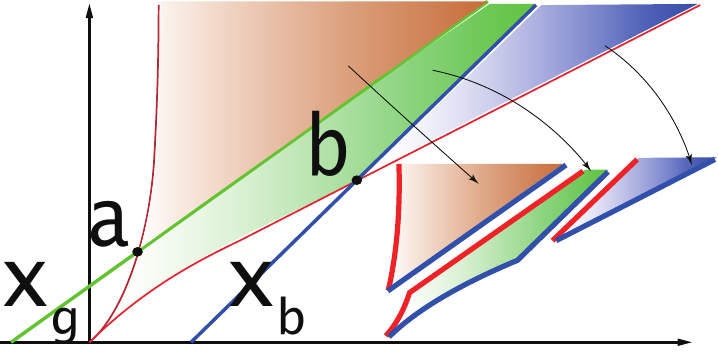}		
		\label{profiles_generation_f}
	\end{minipage}}
	\caption{Examples of different trajectories splitting the \textit{base profile} in side view. Each piece of split profiles is rescaled for illustration. The red and blue lines are new boundaries of each new profile.}
	\label{profiles_generation}

\end{figure}

Currently, we consider TPs in at most three lanes, namely left, current and right lanes. Firstly, the \textit{base profile} $ P_{b} $ is placed in each lane. In Algorithm \algref{alg:3D_algorithm}{alg:3D:core-generate-profiles}, the specially sorted trajectories in each lane split $ P_{b} $ in sequence. Fig.\ref{profiles_generation} lists five basic ways to split the \textit{base profile} by only one trajectory. Taking Fig.\ref{profiles_generation_b} for example, the green trajectory of the rear vehicle intersects the rear boundary of $ P_{b} $ at crunode $ a $. The split derives two TPs and the brown gradient one above $ \mathbf{x}_{g} $ is reserved for connection step. But the green gradient one below $ \mathbf{x}_{g} $ becomes new \textit{base profile} $ P_{b} $. If no trajectory exists any more, then new $ P_{b} $ is reserved for connection too. Otherwise, as shown in Fig.\ref{profiles_generation_f} with a blue trajectory $ \mathbf{x}_{b} $ of the front vehicle, it splits $ P_{b} $ into two new profiles again. In other situations, either a slow front vehicle or a fast rear vehicle will generate some narrow TPs in Fig.\ref{profiles_generation_c}, \ref{profiles_generation_d} and \ref{profiles_generation_e}. If the agent is exactly in the lane, three blue gradient TPs are \textit{root profile} (where our agent locates in) $ P_{ego} $. Under these circumstances, the agent has limited planning space, thus changing to side lanes is in priority. If three narrow TPs are in side lanes, there are eliminated in Algorithm \algref{alg:3D_algorithm}{alg:3D:eliminate_profile}. The elimination benefits decreasing computation time of useless and unsafe routes.

Each TP represents a terminal region in spatio-temporal space, and a specific connection of TPs generates a unique topological route. Algorithm \algref{alg:3D_algorithm}{alg:3D:core-connect-profiles} connects TPs in all combinations with no more than three profiles in a sequence. In other words, the maximum depth of TPs is limited to three, since a route with more than three TPs mostly vanishes quickly in the dynamic environment. Besides, one more connection between TPs means a lane change maneuver with more uncertainty. So a route performing several lane change maneuvers is suppressed. Initially, the agent locates in a lane-keeping route of $ P_{ego} $, and the depth of route equals 1. At searching step, the algorithm checks the overlapping area between $ P_{ego} $ and TPs in adjacent lanes. If a profile $ P_{i}^{j} $ has common projection area with $ P_{ego} $, a new route that ends in $ P_{i}^{j} $ comes into being. Fig.\ref{Dynamic:basic} shows a typical example for the connection searching in Algorithm \algref{alg:3D_algorithm}{alg:3D:connect-overlapping}. $ P_{ego} $ in brown gradient connects to both TPs filled in mesh and dots respectively. Therefore, two left lane-change routes exit. The route of mesh-filled TP leads the agent to the rear left of the green vehicle, while the route that ends in dot-filled TP drives the agent to the front of the green vehicle. At the end of algorithm, all possible routes are enumerated.

\subsubsection{Maneuver window decision}
Although the routes in spatio-temporal space above in different topologies have been found. The exact time instance for a maneuver has not been decided yet. For a lane keeping situation, the agent need no special maneuver. But for a left or right lane change, when and where to perform the maneuver should come into being with the route synchronously. A simple lane-change includes time intervals for beginning and finishing the maneuver. We define the time interval of each maneuver as a \textit{maneuver window} $ W^{i} = (Br_{i},Bl_{i}, Fr_{i}, Fl_{i}), i \in\{1, 2\}$, where $ Br_{i} $ is the more recent time limit for beginning the $ i' $th maneuver, and $ Fl $ is the more later time limit for finishing a maneuver. Besides, the time $ Tb_{i} $ for beginning a maneuver and time $ Tf_{i} $ for finishing a maneuver satisfy 
{\setlength\abovedisplayskip{3pt}
\setlength\belowdisplayskip{1pt}
\begin{equation}
\begin{aligned}
Br_{i}\leq  Tb_{i} \leq  Bl_{i} \\
Fr_{i}\leq  Tf_{i} \leq Fl_{i} \\
Tb_{i} + Te_{i} = Tf_{i} \label{eq:maneuver_window}
\end{aligned}
\end{equation}
}where $ Te_{i} $ is the total time for executing the maneuver.

\begin{figure}
	\centering
	\subfigure[Stereo view]{	\begin{minipage}[t]{0.95\linewidth}
		\includegraphics[width=1\columnwidth, height=0.45\columnwidth]{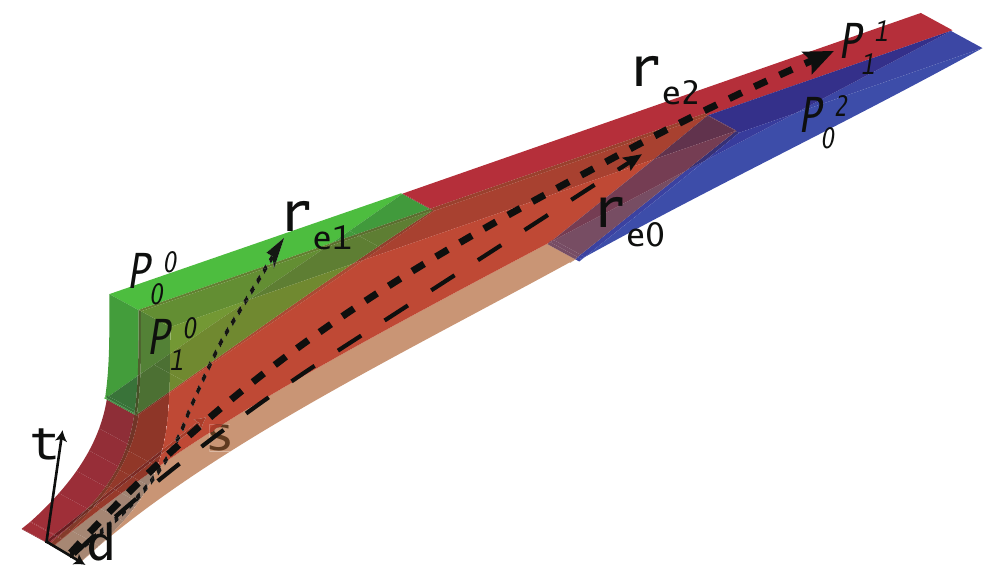}
		\label{Maneuver-window_3D}
		\end{minipage}}
	\subfigure[Side view]{	\begin{minipage}[t]{1.\linewidth}
		\includegraphics[width=1\columnwidth]{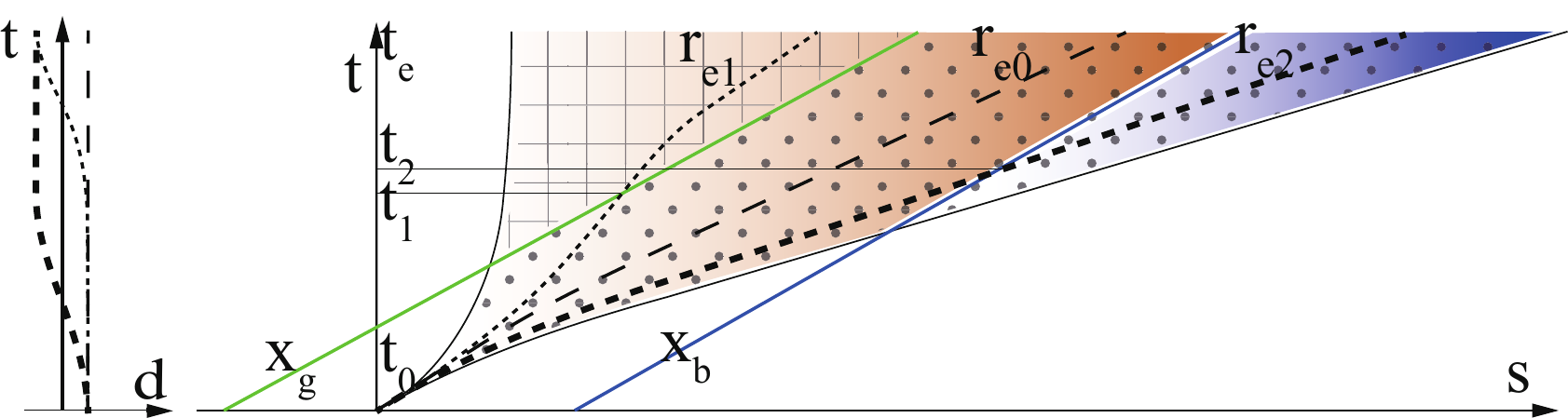}
		\end{minipage}}
	\caption{\myadd{A recurrence of Fig.}\ref{Dynamic:basic} \myadd{in different views. Three dash lines are the potential routes of the maneuvers in different homotopy classes.}}
	\label{Maneuver-window}
\end{figure}

Fig.\ref{Maneuver-window} explains how the \textit{maneuver window} arises. In the stereo view, a potential route $ \mathbf{r}_{e1} $ leads the agent to the back of the green vehicle. Apparently, the agent can only do left change after time $ t_{1} $ at the crunode of $ \mathbf{r}_{e1} $ and $ \mathbf{x}_{g} $ in side view. So the \textit{maneuver window} of $ r_{e1} $ is $ W_{e1}^{1} = (t_{1}, t_{e}, t_{1}, t_{e}) $. Different from $ \mathbf{r}_{e1} $, the route $ \mathbf{r}_{e2} $ has plenty time window to begin a lane change, and it should execute before time $ t_{2} $ at the crunode of $ \mathbf{r}_{e2} $ and $ \mathbf{x}_{g} $. Obviously, the maneuver should finish before time $ t_{2} $ \myadd{too.}\myremove{And}The corresponding \textit{maneuver window} of $ \mathbf{r}_{e2} $ is $ W_{e2} = (t_{0}, t_{2}, t_{0}, t_{2}) $.

Different from traditional sampling based methods, our routes\myremove{generated}are not constrained by fixed discretization time interval. Thus the consistency in routes is guaranteed. 

\subsection{Group Corridors and Routes}
A corridor in $ d $\nobreakdash--$ s $ plane constrains the lateral width along longitudinal direction. While a route in $ s $\nobreakdash--$ t $ plane constrains the longitudinal pose at different time instances with loose constraints by lane. By grouping a corridor and a route involving the same lanes, a specific maneuver is generated with complete boundary constraints. The brown route $ r $ (see Section \ref{opt:longitudinal}) to the left lane has a \textit{maneuver window} $ W^{1} = (t_{0},t_{e}, t_{0}, t_{e}) $ in Fig.\ref{lat_guess}. Then, the \textit{maneuver window} is further truncated by the limitation of effective length of the right lane. As the route $ r $ would collision with red obstacles at time $ t_{1} $, so the latest time of beginning the maneuver is $ t_{1} $. A reasonable maneuver should end in the left lane before $ t_{1} $. Now, the updated \textit{maneuver window} is $ W^{1} = (t_{0},t_{1}, t_{0}, t_{1}) $. 

The meaning of grouping is significant.\myremove{By}\myadd{After}this step, maneuvers are really generated in spatio-temporal space. The boundary, \textit{maneuver window} and involved lanes are integrated together as the basic information for a maneuver. Besides, the maneuvers can be evaluated according to gathered information. For instance, when a route $ r $ has limited \textit{maneuver window}, we prefer to ignore the route in advance for the safety and efficiency sake. With the constraints of \myadd{a}maneuver, the trajectory can be generated with more heuristic information.

\section{Trajectory Optimization}
\label{sec_trajectory_optimizaiont}
Trajectory generation must consider vehicle kinematics and dynamics within the boundaries of corridors. To represent the continuous change of vehicle states, the numerical method is used to generate a trajectory. However, due to the non-convex property of TPs, numeric optimization methods cannot solve the problem of trajectory planning directly utilizing one cost function. Inspired by Schulz\textquotesingle s work \cite{schulz2017EstimationCollectiveManeuvers}, the problem of trajectory planning are decomposed into longitudinal and lateral sub-problems. The weights of each term in the objective function are adjusted to constrain the final solution instead of adding hard constraints on states to be optimized. 

In the end profile of a route, a terminal state is set as the target to optimize the route in $ s $\nobreakdash--$ t $ plane. Then using the boundary and \textit{maneuver window} to limit lateral range of the route along longitudinal direction, an initial guess of lateral pose within both boundaries in predicting time can be derived as show in Fig.\ref{lat_guess}. Eventually, the final trajectory is generated after once more lateral optimization in $ d $\nobreakdash--$ s $ plane.

\subsection{Longitudinal Optimization}
\subsubsection{Longitudinal initial guess}
\begin{figure}[ht]
	\centering
	\includegraphics[width=1\columnwidth]{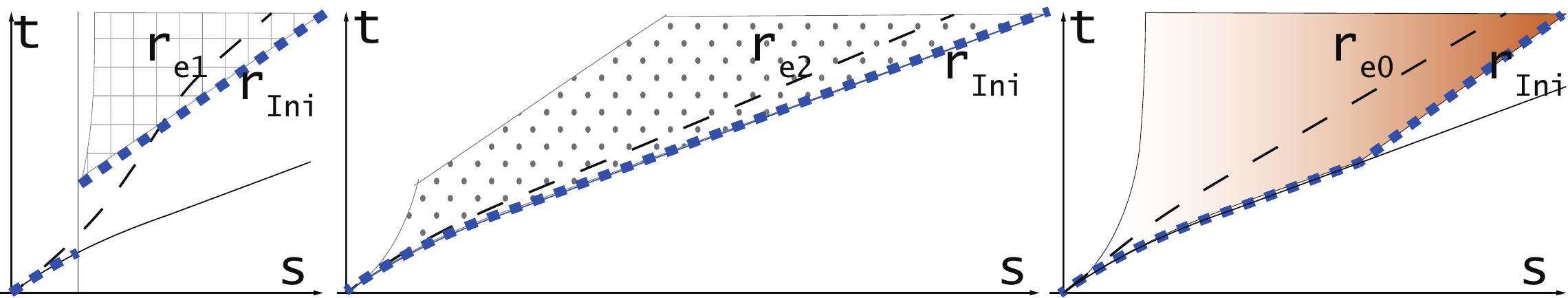}
	\caption{Initial trajectories of Fig.\ref{Dynamic:basic}. The blue dots of $ \mathbf{r}_{Ini} $ are initialized values of three trajectories.}
	\label{Initial-guess}
\end{figure}
\label{opt:longitudinal}
TPs are represented by combining different trajectories of traffic participants and the mobility of the agent. Then, we initialize a guess of a route $ \mathbf{\bar{r}} $ by part of TPs. As shown in Fig.\ref{Initial-guess}, the green dots represent our initialization guess of three\myremove{possible}trajectories of the agent, which are $ \mathbf{r}_{e0} $, $ \mathbf{r}_{e1} $ and $ \mathbf{r}_{e1} $ respectively. Here, we define an initial trajectory as $ \mathbf{\bar{r}}^{N} = \{ \bar{s}_{0},\dots,\bar{s}_{N}\} $ with time-step $\Delta T $, where $ N $ is the the predicting horizon.

\subsubsection{Speed limit}
\begin{figure}[!ht]
	\centering
	\includegraphics[width=0.6\columnwidth]{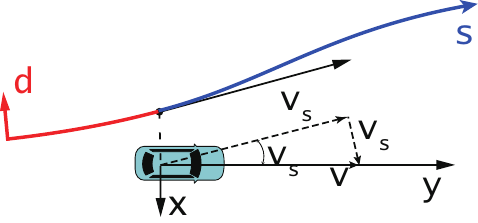}
	\caption{The velocity components in $ s $ and $ d $ direction in Curvilinear Coordinate}
	\label{velocity_decompose}
\end{figure}

As described in Fig.\ref{velocity_decompose}, under the decomposed condition in longitudinal and lateral direction, $ v_{s} $  and $ v_{d} $ are the longitudinal and lateral velocity components respectively. Speed limit refers to the velocity component in $ s $ axis. The physical limit of vehicle constrains the optimization space. The main physical limit comes from the centripetal acceleration. Considering optimizing in longitudinal direction, the velocity is limited by maximum acceleration $ a_{cen} $ and the road curvature $ \rho $ as

{\setlength\abovedisplayskip{3pt}
\setlength\belowdisplayskip{1pt}
\begin{equation}
v_{acc} = \sqrt{\frac{a_{cen}}{\rho} }\label{eq:v_limit_by_curvature}
\end{equation}}Besides, the speed limit $ v_{sig} $ of the road signs or traffic lights should also be considered. Under these constraints, the velocity at predicting time can be initialized as 
{\setlength\abovedisplayskip{3pt}
	\setlength\belowdisplayskip{1pt}
\begin{equation}
 \bar{v}_{N} = \min(v_{acc}, \max(v_{sig}, \bar{v}-N*a_{dec}))  	\label{eq:end_v_limit_min}	
\end{equation}}where $ \bar{v} $ is the current velocity and $ a_{dec} $ is the deceleration speed.

\subsubsection{End pose limit}
The trajectory should keep some distance with the front vehicles both in terms of safety and extra maneuver space. Especially, when the agent drives fast, it often has a limited time or space to perform a maneuver when an urgent situation happens. The total delay-time $ t_{delay} $ for perception, decision and brake-lag approximates 1.5 seconds in our system.\myremove{Besides, some extra space $ L_{Extra} $ and vehicle length $ L_{v} $ for safety are added.}\myadd{For the passengers' feeling and social acceptance, some extra space $ L_{Extra} $ is considered as a function of ego velocity, and $ L_{Extra} $ is proportional to ego vehicle's velocity:}
\begin{equation}
\myadd{L_{Extra}=\alpha_{Extra}*\bar{v} + L_{Extra\_min}}
\end{equation}
\myadd{where $ \alpha_{Extra} $ is the coefficient parameter for velocity determined empirically and $ L_{Extra\_min} $ is the minimal extra distance at zero velocity. Besides, vehicle length $ L_v $ is added for safety.}Other factors such as road fluctuation, weather condition and the weight of vehicles are not considered.
Then, the end pose in $ s $ axis is initialized by 
{\setlength\abovedisplayskip{3pt}
\setlength\belowdisplayskip{3pt}
\begin{equation}
\bar{s}_{N} \leq s_{N} -(\bar{v} t_{delay} + L_{Extra} + L_{v})\label{eq:end_s_limit}
\end{equation}}where $ s_{N} $ is the pose in the end TP of a maneuver.

\subsubsection{Cost function}
For the continuous control of the vehicle and the comfort of passengers, we select a quadratic cost-function to penalize the uneasy acceleration and the deviation from either the end pose $ \bar{s}_{N} $ or the end velocity $ \bar{v}_{N} $. The \textit{longitudinal cost} is defined as:
{\setlength\abovedisplayskip{1pt}	\setlength\belowdisplayskip{1pt}
\begin{align}\label{eq:j_s}
\mathbf{J}_{s} = \sum_{i = 0}^{N}( \omega_{i,a}^{s}j_{i,a}^{s} + \omega_{i,jk}^{s}j_{i,jk}^{s} )\notag\\
  + (\omega_{0,v}^{s}j_{0,v}^{s} + \omega_{N,v}^{s}j_{N,v}^{s})\\
   + (\omega_{0,p}^{s} j_{0,p}^{s} + \omega_{N,p}^{s}j_{N,p}^{s})\notag 
\end{align}}The influence of each term can be tuned with the weight factors $ \omega_{i,a}^{s} $, $ \omega_{i,jk}^{s} $, etc. To constrain the start pose and to lead the agent from the current pose to the target pose, the \textit{position cost} $ j_{i,p} $ is defined as
\begin{equation}
	j_{i,p}^{s} = (s_{i} - \bar{s}_{i})^{2}, i\in\{0,N\} \label{eq:j_s_ini}
\end{equation}
The value of cost increases when the optimized route deviates from either the target end pose or the current pose. To make the start and end velocity as close as possible to the given values $ \bar{v}_{0}^{s} $ ( equals $ \bar{v} $) and $ \bar{v}_{N}^s{} $ respectively, the second term, the \textit{velocity cost} $ j_{i,v} $  is defined as 
{\setlength\abovedisplayskip{3pt}
	\setlength\belowdisplayskip{1pt}
\begin{equation}
	j_{i,v}^{s} = (v_{i}^{s} - \bar{v}_{i}^{s})^{2}, i\in\{0,N\} \label{eq:j_s_velocity}
\end{equation}}To make the agent move smoothly and avoid jerky action, the first two terms \textit{acceleration cost} $ j_{a}^{s} $ and \textit{jerk cost} $ j_{jk}^{s} $ are defined as
\begin{subequations}\label{eq:j_s_acc_jerk}
	\begin{align}
		j_{i,a}^{s} = a_{s,i}^{2}\\
		j_{i,jk}^{s} = j_{s,i}^{2}
	\end{align}
\end{subequations}

By using forward difference, the values of velocity, acceleration and jerk are approximated as 
\begin{subequations}\label{eq:v_a_j_finite_difference}
\begin{align}
	v_{s,i} &= \frac{ s_{i+1}-s_{i} }{ \Delta T }\\
	a_{s,i} &= \frac{ s_{i+2}-2s_{i+1}+s_{i} }{ (\Delta T)^{2} }\\
	j_{s,i} &= \frac{ s_{i+3}-3s_{i+2}+3s_{i+1}-s_{i} }{ (\Delta T)^{3} }
\end{align}
\end{subequations}

\subsection{Lateral Optimization}
The step-wise optimization is the key step for optimizing our trajectories. Without the prior knowledge of the optimized route in $ s $\nobreakdash--$ t $ plane, directly solving trajectory planning in spatio-temporal space is quite difficult. One choice is to solve this problem with non-convex constraints as an MIP (mixed-integer programming) problem \cite{park2015HomotopyBasedDivideandConquerStrategy}. The other is to evaluate vast paths sampled in spatio-temporal space. However, both methods demand for a cost of time or computation load (or both).

With longitudinal route generated first, we can get the initial guess for the lateral poses. Taking Fig.\ref{lat_guess} for example, the shorter blue route only involves the right lane. The lateral initial guesses are limited by boundaries in green dash lines. As for the longer brown route leading the agent to the left lane, it covers two lanes passing the obstacles in the right lane. Thus, the initial guesses should be constrained between brown dash lines. Apparently, with the green route generated in $ s $\nobreakdash--$ t $ plane, the initial lateral pose  should be limited in the narrower left lane after time $ t_{1} $.
\subsubsection{Lateral initial guess}
\begin{figure}[!ht]
	\centering
	\subfigure[Boundaries of two corridors. The thin dash lines are corresponding boundaries of a route in the same color. The thick dash lines are the initial lateral guess of brown route.]{	\begin{minipage}[t]{0.85\linewidth}
		\includegraphics[width=1\columnwidth]{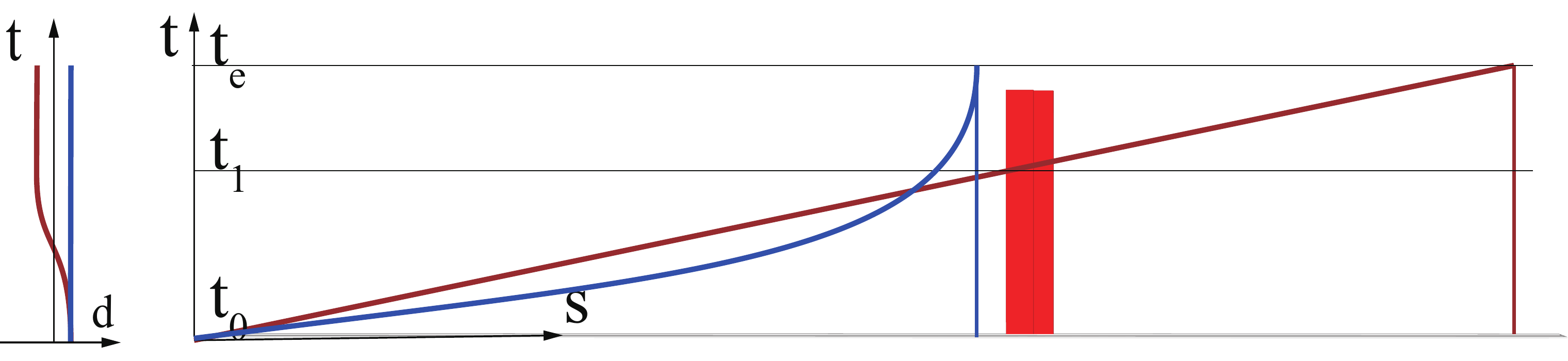}
		\label{lat_guess_side}
	\end{minipage}}
	\subfigure[\myadd{maneuver window of two routes in left and side views.}]{ \begin{minipage}[t]{0.85\linewidth}
		\includegraphics[width=1\columnwidth]{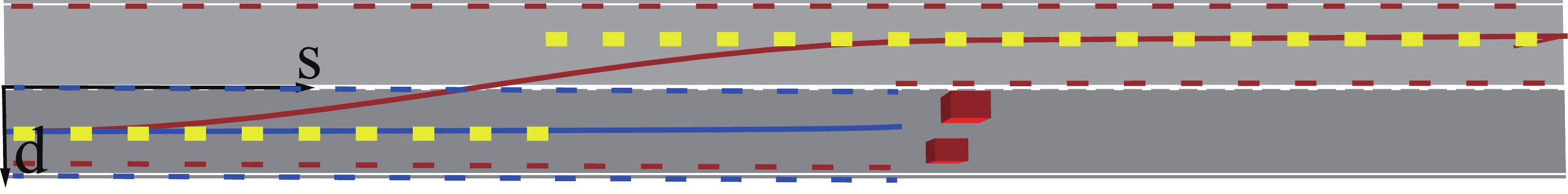}
		\label{lat_guess_bird}
	\end{minipage}}
	\caption{\myadd{Example of lateral initial guesses of two trajectories in different homotopy paths.}}
	\label{lat_guess}
\end{figure}

In this paper, we select a lane center as the initial lateral guess according to \textit{maneuver window} first. Taking the brown route in Fig.\ref{lat_guess} for example, the lateral initial guess includes two parts. The centerline of the right lane from time $ Tb_{1} $ to $ Tb_{1}+ Te_{1}/2 $ is the first part, and the centerline of the left lane from time $ Tb_{1}+Te_{1}/2 $ to $ Tf_{1} $ is the second part. The initial guess of a route is adjusted by checking whether the guess is in the boundaries of the maneuver\textquotesingle s corridor. If the guess is too close to or out of boundary, we nudge the guess toward the center of two boundaries. Therefore, we have the lateral initial guesses as
\begin{align}
\bar{d}_{i}= 
\begin{cases}
d_{i}^{low}+d_{safe}^{prj},& \text{if } \bar{d}_{i}\leq d_{i}^{low}+d_{safe}^{prj}\\
d_{i}^{up}-d_{safe}^{prj},& \text{if }  \bar{d}_{i}\geq d_{i}^{up}-d_{safe}^{prj}
\end{cases}
\end{align}where $ d_{i}^{low} $ and $ d_{i}^{up} $ are the values of the lower and upper boundaries respectively, $ d_{safe}^{prj} $ is the projected length  of $ d_{safe} $ in the lateral direction.

\subsubsection{Cost function}
The lateral poses are optimized in a manner similar to that in the longitudinal direction. The only difference lies in the \textit{position cost} and the \textit{velocity cost}. The optimized lateral poses should transfer to the initial guess smoothly. Firstly, the lateral poses should not deviate too far from the initial guess. Secondly, lateral change should be as smooth as possible in the range of \textit{maneuver window}. Thus, we consider the \textit{position cost} and \textit{velocity cost} at every time instance. Finally, the \textit{lateral cost} is defined as 
{\setlength\abovedisplayskip{3pt}
	\setlength\belowdisplayskip{1pt}
\begin{align}
	\mathbf{J}_{d} = \sum_{i=0}^{N}(\omega_{i,a}^{d}j_{i,a}^{d}+\omega_{i,jk}^{d}j_{i,jk}^{d}+\omega_{i,p}j_{i,p}^{d}+ \omega_{i,v}^{d}j_{i,v}^{d}) \label{eq:j_d}
\end{align}}where \textit{position cost} $ j_{p}^{d} $, \textit{velocity cost} $ j_{v}^{d} $, \textit{acceleration cost} $ j_{a}^{d} $ and \textit{jerk cost} $ j_{jk}^{d} $ are defined in the similar form of longitudinal ones in equations \eqref{eq:j_s_ini}, \eqref{eq:j_s_velocity}, \eqref{eq:j_s_acc_jerk} and \eqref{eq:v_a_j_finite_difference}. We increase the weights of \textit{position cost} at the time instance where a great change of lateral pose happens. The weights of \textit{velocity cost} $ \omega_{i,v}^{d} $ are different in every loop. $ \omega_{i,v}^{d} $ increases linearly with the square of the lateral offset to target position. 
{\setlength\abovedisplayskip{3pt}
	\setlength\belowdisplayskip{1pt}
	\begin{equation}
		\omega_{i,v}^{d}\propto d_{0}^{2} 
	\end{equation}
}When the agent gets away from the target lane center, $ j_{p}^{d} $ increases linearly to the square of the shift distance in lateral direction and overwhelms other costs. So, We increase the weights of $ j_{v}^{d} $ to balance the costs and suppress lateral velocity in the meanwhile.

\subsection{Numeric Optimization}
Optimization-based methods have been vastly used for trajectory planning problem\cite{zieglerTrajectoryPlanningBertha2014a, park2015HomotopyBasedDivideandConquerStrategy, rosmann2017IntegratedOnlineTrajectorya, lim2018HierarchicalTrajectoryPlanning}. Combining the advantages of both Gauss-Newton method and steepest descent method, LM(Levenber-Marquradt) algorithm can find a solution even if LM starts far off the real solution. Although LM can be a little slower than Gauss-Newton method, the cost of slow speed is negligible within our problem. To find out the solution of equation \eqref{eq:j_s} and \eqref{eq:j_d}, we use Ceres solver to carry out LM algorithm. Ceres has been well tested and is agile enough to adjust residual, weights and other parameters.  

\begin{table*}[!ht]
	\caption{{ Comparison of several state-of-the-art Approaches considering maneuver decisions and planning.}}
	\label{T:table_compare}
	\centering
	\begin{tabular}{@{}l| l l l l l l l l}
		\hline
		\myadd{No\#}&\myadd{Method} &\myadd{PH} 	&\myadd{Global Strategy} 	&\myadd{Optimization} &\myadd{ST R(mxm/s)} &\myadd{TA} &\myadd{$ \nabla T $} & \myadd{CT}\\ \hline
		
		\myadd{A1}&\myadd{Gu et al}\cite{gu2016AutomatedTacticalManeuver,gu2017ImprovedTrajectoryPlanning}\ & \myadd{6s }
		&\myadd{DAG,\ sampling }&\myadd{iLQR} &\myadd{(4x0.4)/2} &\myadd{\checkmark}  & \myadd{0.1s} &\myadd{\textminus} \\
		
		\myadd{A2}&\myadd{S\"{o}ntges et al}\cite{sontges2018ComputingDrivableArea} &\myadd{3s} & \myadd{RSP} &\myadd{via sampling} &\myadd{(0.4x0.4)/0.15} &\textminus &\myadd{0.15s} &\myadd{75ms}\\
		
		\myadd{A3}&\myadd{Lim et al}\cite{lim2018HierarchicalTrajectoryPlanning} &\myadd{3s} &\myadd{Hierarchical} &\myadd{A*,SQP} &\myadd{(5x\textminus)/0.5} &\myadd{\textminus} &\myadd{0.1s}  &\myadd{100ms} \\
		
		\myadd{A4}&\myadd{Ziegler et al}\cite{zieglerTrajectoryPlanningBertha2014a} &\myadd{10s} & \myadd{Polygonal split} &\myadd{SQP} &\myadd{$ \forall $} &\myadd{\textminus} &\myadd{0.33s } &\myadd{\textminus} \\
		
		\myadd{A5}&\myadd{Park et al}\cite{park2015HomotopyBasedDivideandConquerStrategy} &\myadd{3s} &\myadd{Cell Sequences} &\myadd{MPC,MIQP} &\myadd{$ \forall $} &\myadd{\checkmark} &\myadd{0.15s} &\myadd{248ms}\\
		
		\myadd{--}&\myadd{Proposed} &\myadd{10s} &\myadd{Enumerate TPs } &\myadd{QP} & \myadd{$ \forall $} &\myadd{\checkmark} &\myadd{0.25s}  &\myadd{38ms}\\
		\hline
	\end{tabular}
	\begin{flushleft}
		{{ { Abbreviations: PH: Planning horizon, STR: Spatio-temporal resolution, TA: Topological awareness, $ \nabla T $: Sampling time, CT: Computation time, \textminus: not given or not considered, $ \forall $: the resolution is dependent on requirements(it can be an arbitrary value), DAG:Directed acyclic graph, iLQR: iterative Linear Quadratic Regulator, RSP: Reachable set propagation, QP: Quadratic programming optimization, MPC: Model predict control, MIQP: Mixed-integer quadratic programming, SQP: Sequential quadratic programming, TPs: Trajectory Profiles}} }
	\end{flushleft}
\end{table*}

Two chief factors contribute to optimizing several trajectories at the same time. One is the heuristic information generated from the profile and the lateral constraints from the corridor\myremove{in}\myadd{of}a given maneuver. The other are the adjusted weights of\myremove{cost-terms in the cost function}\myadd{cost functions.}The former initializes parameters to be optimized with high accuracy, while the latter avoids hard parameter constraints. Both of them can ensure\myremove{our}trajectory optimization to be a real-time one and guarantee a feasible solution.

\section{Experimental Evaluation}
\label{sec_exp}
Below we evaluate the proposed SMSTP algorithm\footnote{{See }\url{https://youtu.be/K2HdANtOjvE}{ for video demos.}}.\myremove{We first test the performance of SMSTP algorithm through some difficult scenarios in a simulation platform. Afterwards, we test some decision-oriented task by considering lane changes in real-world dynamic traffic.}
\myadd{We first compare SMSTP algorithm with the state-of-the-art works. Afterwords we test the performance of the algorithm through emergence scenarios in a simulation platform. Finally, we test some decision-oriented task by considering lane changes in real-word dynamic traffic.}

\subsection{System And Run Time }

Our platform is built on Linux and uses RCS (Real Time Communication) for inter-process communication. For simulation part, the lane lines are manually labelled using ArcGIS and queried in local PostgreSQL database. The traffic participants are randomly generated with varying target speed and behaviours. For real dynamic traffic test, the driving test is implemented on HongQi 3 installed with a lane detecting camera, two long range radars for front view and rear view respectively, a front view 8-line laser and  a middle-range 32-line Lidar on the roof. And lane lines are real-time detected by a forward camera. The cost-map $ m_{cost} $ has $ 250 \times 800 $ cells at 0.2 m resolution. Our algorithm is implemented on Inter Core i7\emph{@}2.40GHz in C++ language.

Table \ref{table_times} lists the running times for the main parts of SMSTP algorithm. Our algorithm is time-saving for real-time autonomous system. The main cost is the time for optimization, and the time may fluctuate within up to 37.1 ms. A critical cause of time rise and fall lies in the varying number of corridors and routes in different environments. However, the traffic situations constrain the number of corridors and routes automatically. When the number of the obstacles increases, road condition is worse. And the overall speed of traffic and the predicting horizon of the agent decreases, thus fewer vehicles run into the \textit{base profile} of the agent. Besides, when the agent is at higher speed, the increasing number of vehicles in the \textit{base profile} results in less spatio-temporal space in each topological space. In that case, some routes with narrow \textit{maneuver window} can be ignored before optimizing. In short, the algorithm has an adaptive ability to balance the time for optimization and the number of feasible alternative solutions.

\begin{table}[!h]
	\caption{Computation times for the main operations in algorithms}
	\label{table_times}
	\centering
	\begin{tabular}{@{}l l l }
		\hline
		\multirow{2}{*}
		{operation} &
		\multicolumn{2}{c}{computation time}\\
		\cline{2-3}
		& average & maximum\\ \hline
		Topology by static & 1.9ms &  4.9ms\\
		Topology by dynamic & 0.2ms & 0.9ms\\
		Total Optimization	& 17.ms & 37.1ms\\
		\hline
	\end{tabular}
\end{table}

\subsection{\myadd{Comparison With state-of-the-art Works}}

\myadd{A comparison of several state-of-the-art approaches considering maneuver decisions and trajectory planning for on-road driving is shown in table \ref{T:table_compare}. The main features are the maximum planning horizon, the global strategy, the optimization method, the STR( Spatio-Temporal Resolution), Topological awareness, sampling time and the computation time. Note that the STR means the minimal difference between two trajectories, and this depends on the sampling distance in three axes of the coordinate. Whereas, the sampling time $ \nabla T $ means the time difference of the same trajectory, and this decides the length of a trajectory.}

\myadd{The sampling based methods, \textbf{A1}, \textbf{A2}  and \textbf{A3} has a lower spatio-temporal resolution, which means these algorithms can hardly get global optimal solutions. Besides, the planning horizon is relatively shorter. The difficulty in harmonizing the contradiction of low spatio-temporal resolution and longer planning horizon lies in enormous spatio-temporal space and limited computation resources. Compared to these methods, SMSTP algorithm plans a longer horizon more quickly and the generated trajectory is not limited by spatio-temporal resolution. This means SMSTP algorithm can reason about the word more predictive and accurately.
\textbf{A2}, \textbf{A3} and \textbf{A4} cannot reason about the environment with topological awareness as they only aim at reducing a single cost. Whereas topological awareness helps maneuver decisions in dealing with semantic information from global task. In fact, \textbf{A5} is a good planner in the aspects of both STR and topological awareness. However, the computation time is quite longer while using MIQP formulation for running linear MPC. For example, it takes 248 milliseconds to optimize a single maneuver with only once lane change. In SMSTP algorithm, the step-wise optimization calculates efficiently with boundary constraints as initial values and a heuristic state as a terminal target.
In short, combining a long planning horizon and topological awareness endows the autonomous vehicle with the ability to maneuver globally but not reactively.
}

\subsection{Simulation Evaluation}
\begin{figure}[!htbp]
	\centering
	\subfigure[\myadd{Profiles and trajectories in left and Side views.}]
	{\begin{minipage}[b]{1.\linewidth}
		\includegraphics[width=0.9\columnwidth,height=2.cm]{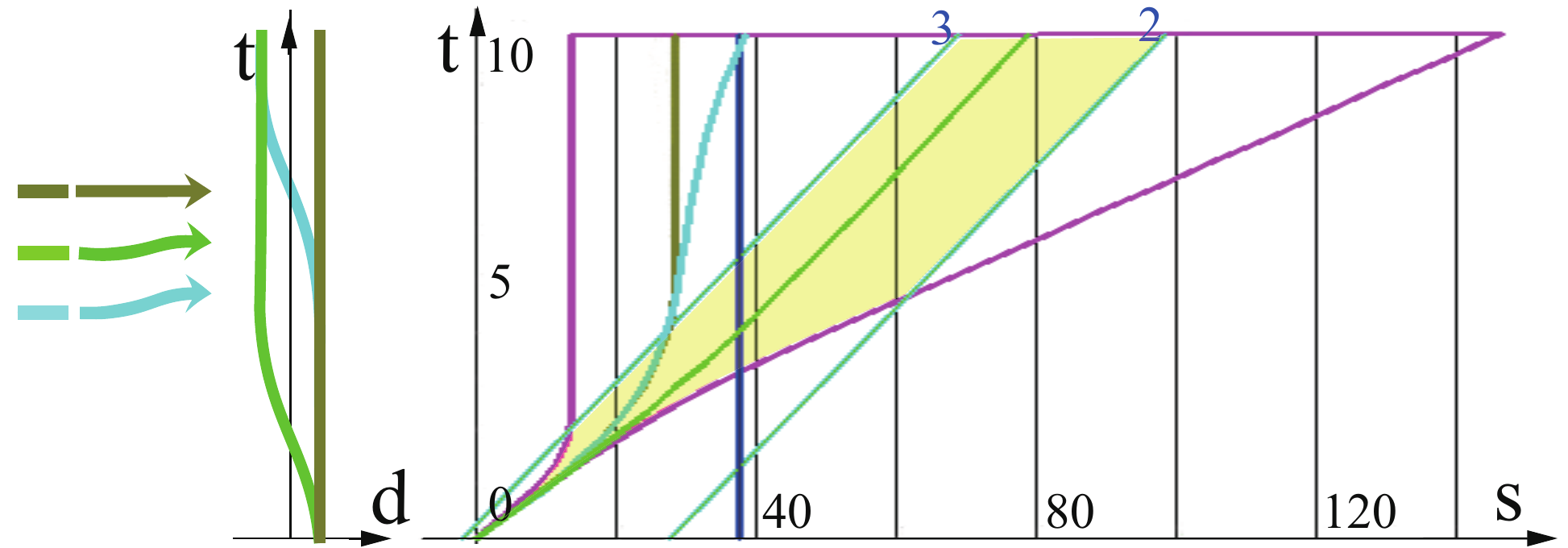}	
		\label{fig:group_vertex}
	\end{minipage}}
		
	\subfigure[Trajectory(green points) in traffic flow. The blue triangles are traffic participants, and the green triangle is the ego agent.]
	{\begin{minipage}[t]{0.9\linewidth}
		\includegraphics[width=1\columnwidth,height=1.cm]{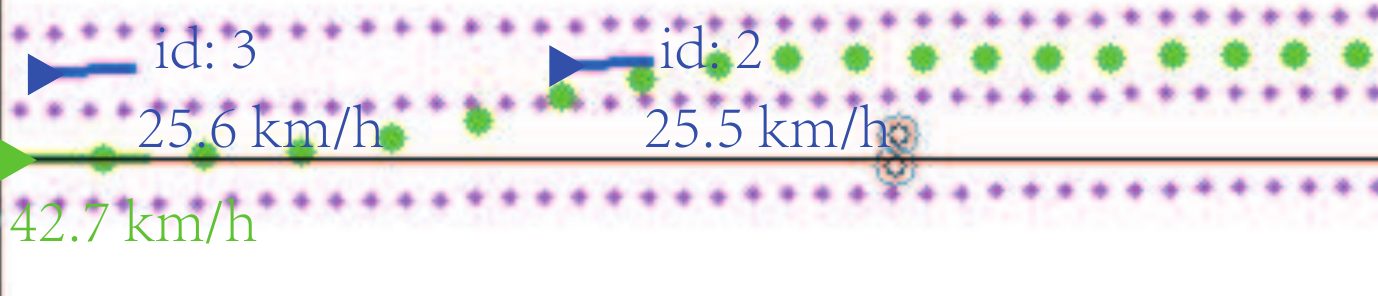}
		\label{fig:group_vehicles}
	\end{minipage}}

	\subfigure[Width of the corridor(blue points) to left lane.]
	{\begin{minipage}[t]{0.9\linewidth}
		\includegraphics[width=1\columnwidth,height=1.cm]{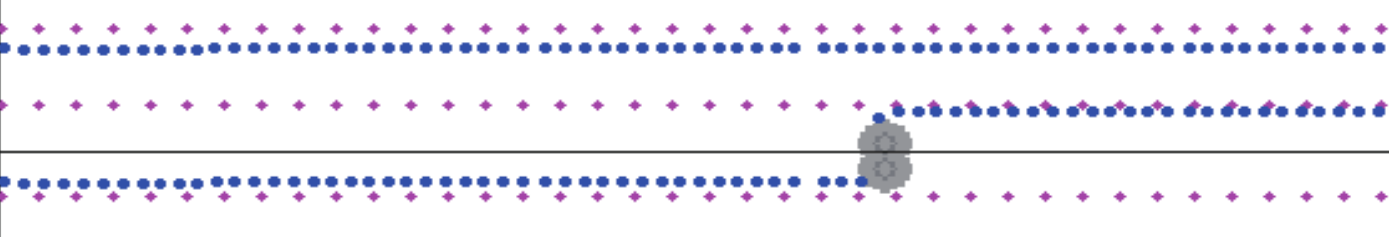}	
		\label{fig:group_static}
	\end{minipage}}
	\caption{Evaluation of the SMSTP algorithm in an emergent situation. The green line  \textquotesingle {\color[rgb]{0,1,0}---}\textquotesingle\  is the selected trajectory. The light yellow profile is the \textit{end profile} of the selected trajectory.}
	\label{fig:group}
\end{figure}

In this section, our SMSTP algorithm is evaluated in the simulation platform on a one-way, two-lane way. We test our algorithm in an emergent situation in Fig.\ref{fig:group}. In the scenario, the obstacles block the right lane, and the agent has to merge into the left lane within limited time or stop as soon as possible. With two near vehicles in the left lane, the initial speed of the further vehicle $ V_{2} $ is 25.5 km/h (7.1 m/s) and the initial speed of the nearer vehicle $ V_{3} $ is 25.8 km/h(7.2 m/s). The trajectories of traffic participants are simply assumed as constant speed, however, any other predicted trajectory can be integrated into the algorithm. The time interval of each point in Fig.\ref{fig:group_vehicles} is 0.5 seconds. The default maximum speed of the agent is 60 km/h, and current speed of the agent is 42.7 km/h (11.9 m/s). So the predicting view in longitudinal direction is about 145 meters.

In this experiment, our aim is to evaluate whether the algorithm is able to generate all possible maneuvers and select a feasible trajectory in emergency. As show in Fig.\ref{fig:group}, three maneuvers are generated. Two maneuvers merges to the left lane, and one maneuver slows down to stop in current lane. Taking the selected green trajectory to the left lane, for example, firstly, a corridor to the left lane is generated and the width of corridor is evaluated as described in Section \ref{Topological_static}. This step avoids collision with obstacle and limits the lateral pose in blue point of Fig.\ref{fig:group_static}. Then, in Section \ref{Topological_dynamic}, a topological route represented by TPs in light yellow to the left lane is generated in Fig.\ref{fig:group_vertex}. Finally, a step-wise optimization get the selected trajectory.

Originally, at least four routes exist. These routes include one route ending in current lane before the obstacle, one route going to the back of $ V_{3} $, one route merging into the place between $ V_{3} $ and $ V_{2} $ and one route $ r_{virtual} $(not shown) going to the front of $ V_{2} $. However, at the step of grouping the corridors and routes, $ r_{virtual} $ is removed as collision with static obstacle happens in the third second before overtaking  $ V_{2} $ finishes in Fig.\ref{fig:group_vertex}. So three feasible trajectories exit to the end. And by evaluating the cost of comfort, safety and efficiency, the green trajectory is selected as the final output.

\begin{figure}[!htbp]
	\centering
	\subfigure[Global view of two consecutive maneuvers.]{\begin{minipage}[t]{1\linewidth}
		\includegraphics[width=1\columnwidth,height=3cm]{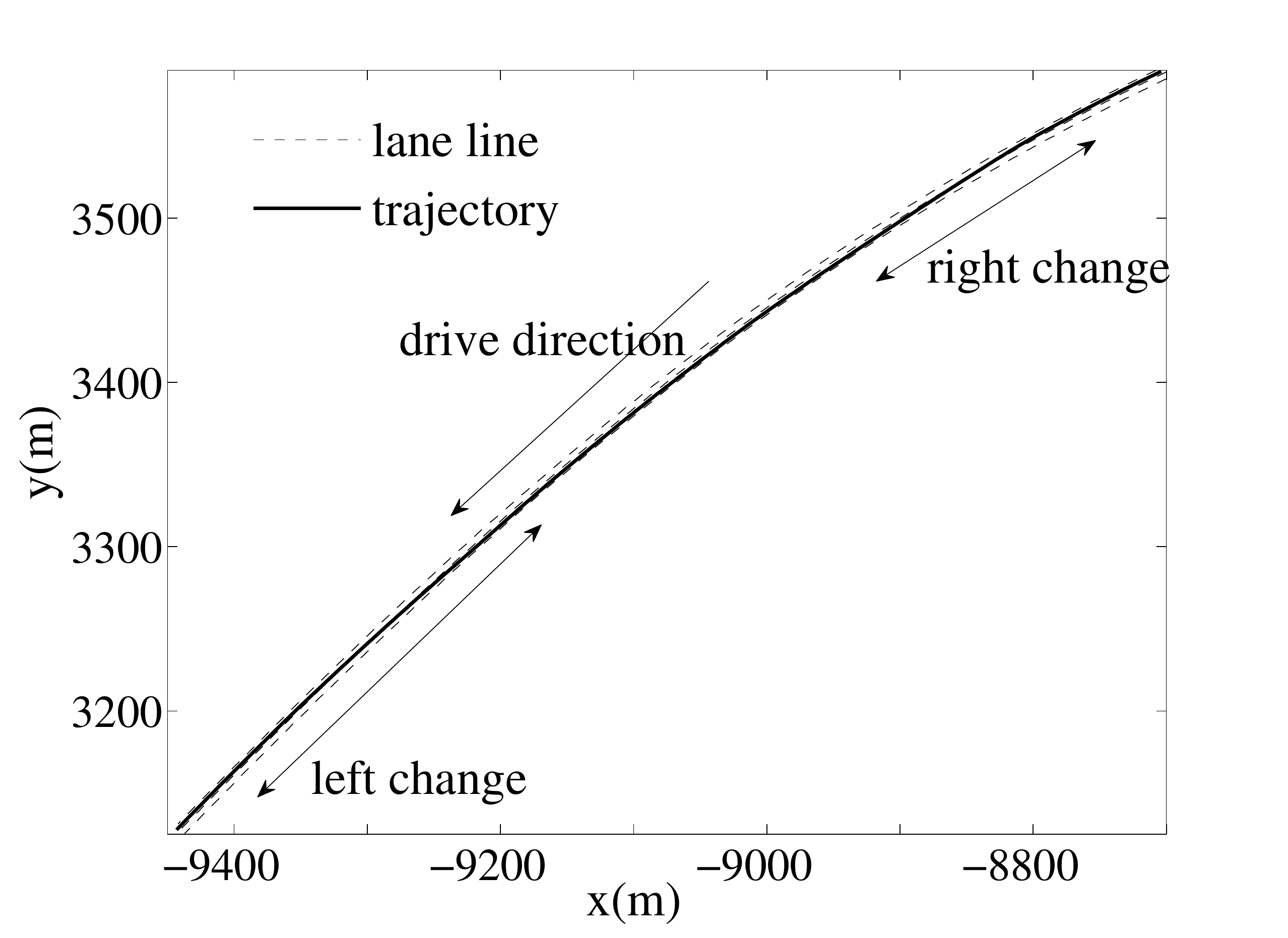}
		\label{exp_gl}
	\end{minipage}}\quad
	\subfigure[Front view of lane keeping.]{\begin{minipage}[t]{0.45\linewidth}
		\includegraphics[width=0.95\columnwidth,height=1.5cm]{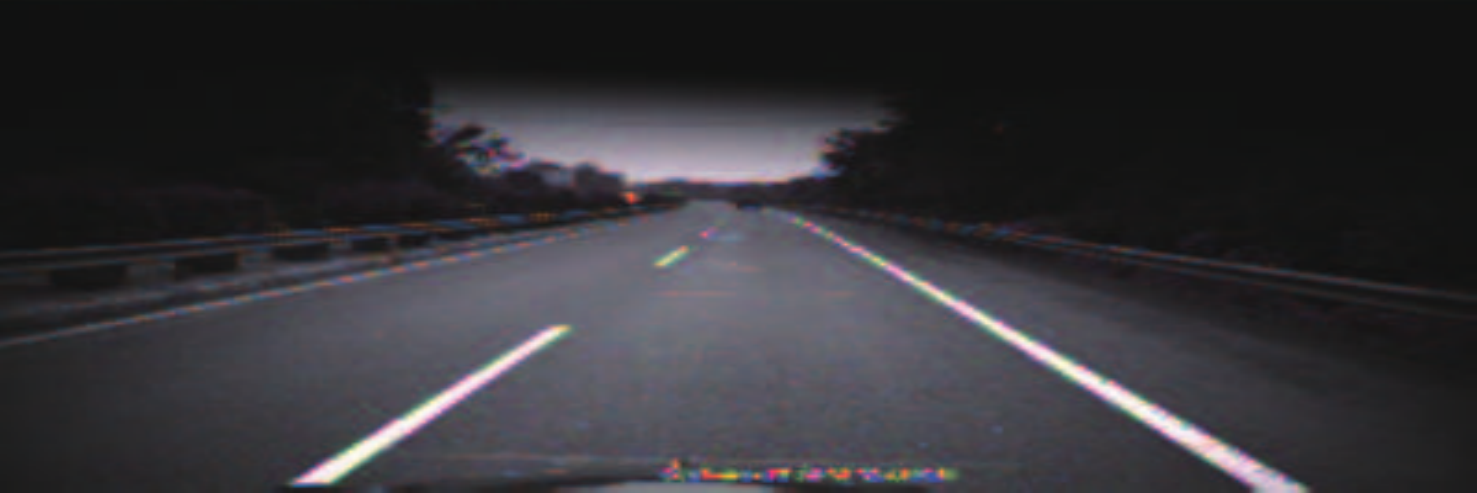}
		\label{exp_lc_f}
	\end{minipage}}
	\subfigure[Front view of left change.]{\begin{minipage}[t]{0.45\linewidth}
		\includegraphics[width=0.95\columnwidth,height=1.5cm]{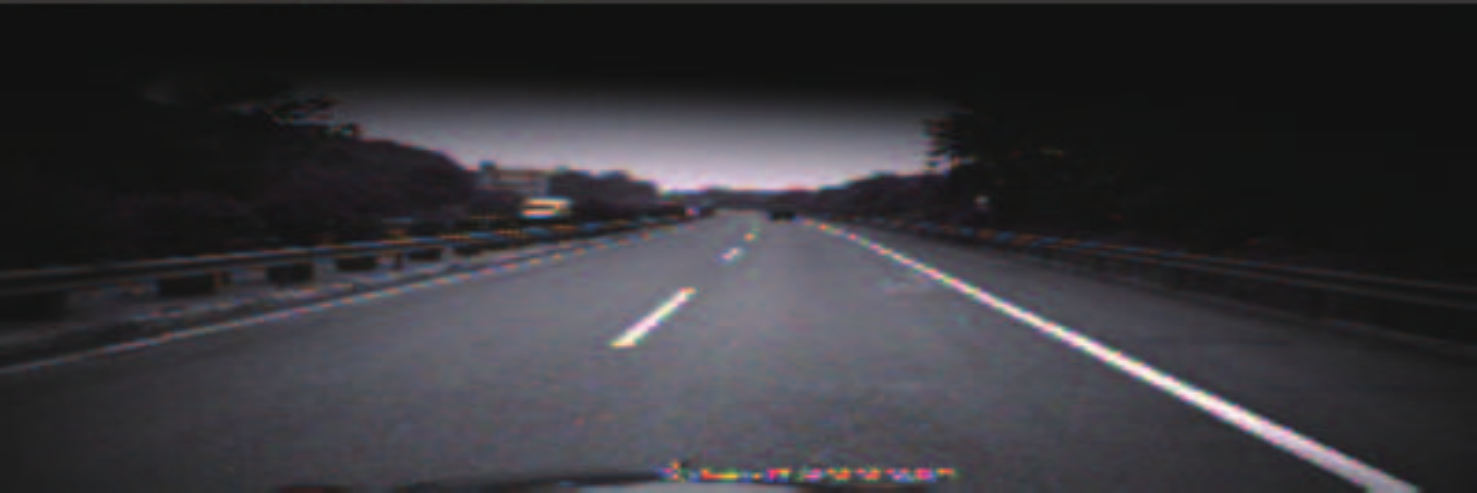} 
		\label{exp_lc2_f}
	\end{minipage}}
	\subfigure[States and planned trajectory of lane keeping.]{\begin{minipage}[t]{0.9\linewidth}
		\includegraphics[width=1\columnwidth,height=0.8cm]{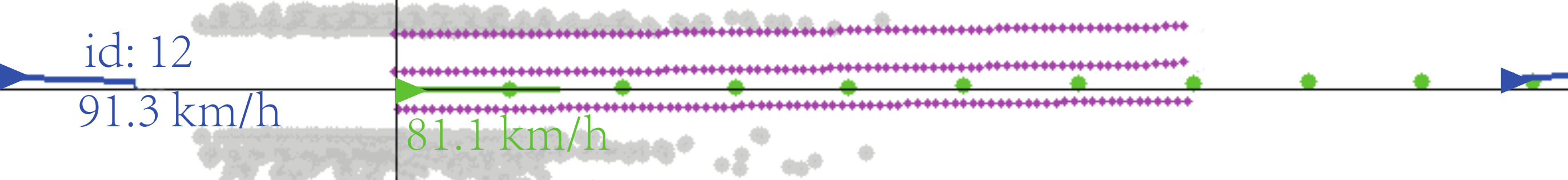} 
		\label{exp_lc_vc}
	\end{minipage}}
	\subfigure[\myadd{Profiles and routes of lane keeping in left and side views.}]{\begin{minipage}[t]{0.9\linewidth}
		\includegraphics[width=1\columnwidth,height=2cm]{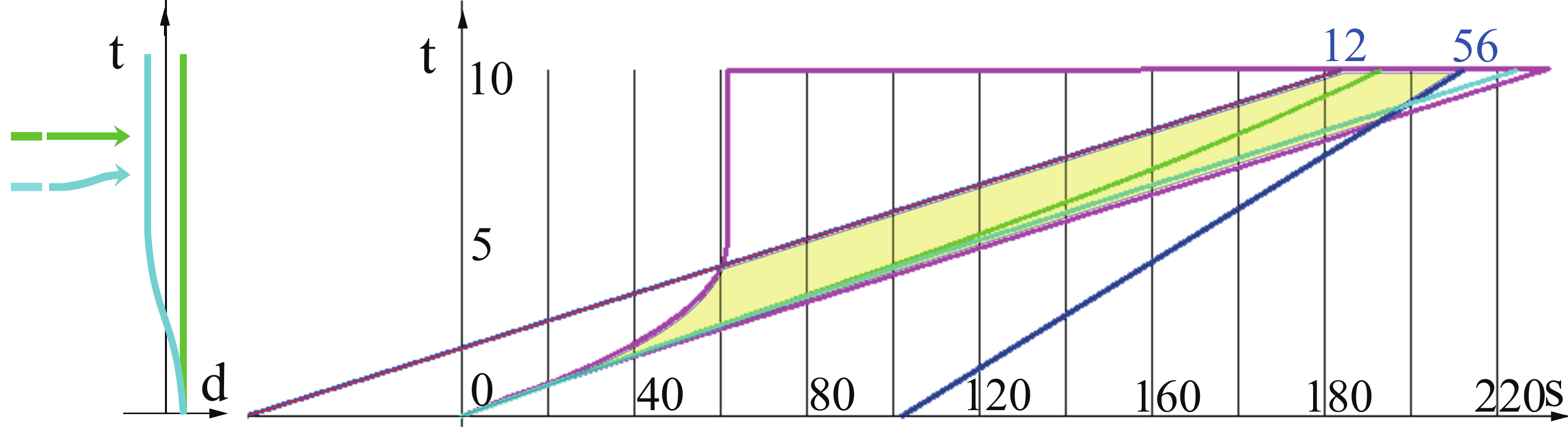}	
		\label{exp_lc_vt}
	\end{minipage}}
	\subfigure[States and planned trajectory of left change.]{\begin{minipage}[t]{0.9\linewidth}
		\includegraphics[width=1\columnwidth,height=0.8cm]{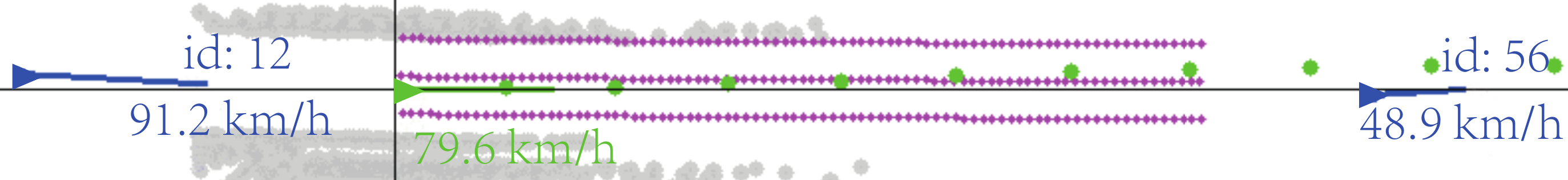}	
		\label{exp_lc2_vc}
	\end{minipage}}
	\subfigure[\myadd{Profiles and routes of left change in left and side views.}]{\begin{minipage}[t]{0.9\linewidth}
		\includegraphics[width=1\columnwidth,height=2cm]{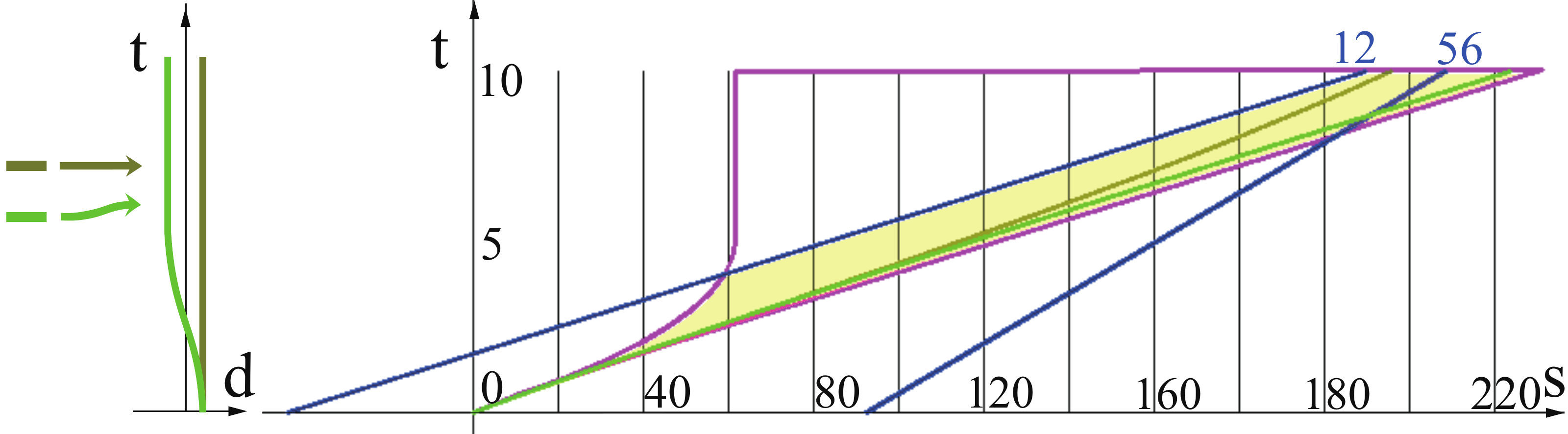}
		\label{exp_lc2_vt}
	\end{minipage}}
	
	\caption{Maneuvers selection in a scenario of overtaking a slow vehicle. Fig.\protect\subref{exp_lc_f},  \protect\subref{exp_lc_vc}, and \protect\subref{exp_lc_vt} illustrate the states of lane keeping. Fig.\protect\subref{exp_lc2_f}, \protect\subref{exp_lc2_vc}, and \protect\subref{exp_lc2_vt} depict the states of left change for overtaking the slow vehicle.}
	\label{exp_all}
\end{figure}

\subsection{Tests in Real-world Dynamic Traffic}
To further study the ability of our SMSTP algorithm, a scenario of overtaking a slow vehicle on the suburban express-way in Changsha City is presented in Fig.\ref{exp_all}.  The agent initially drives in the right lane with a slow vehicle $ V_{56} $ of 48.9 km/h(13.6 m/s) in the front and a fast vehicle $ V_{12} $ of 91.2 km/h(25.3 m/s) in the back. In the experiment, we set the cost function of selecting a faster trajectory among the safe ones and choosing the drive lane (the right one) with priority. Safety means a trajectory without collision to traffic participants and has been existed for some periods with enough maneuver window left. Notice that the topological corridors constraints from static obstacles are removed for this high speed test, since the lane has no obstacles during test. 

\begin{figure}[!htbp]
	\centering
	\includegraphics[width=0.95\columnwidth]{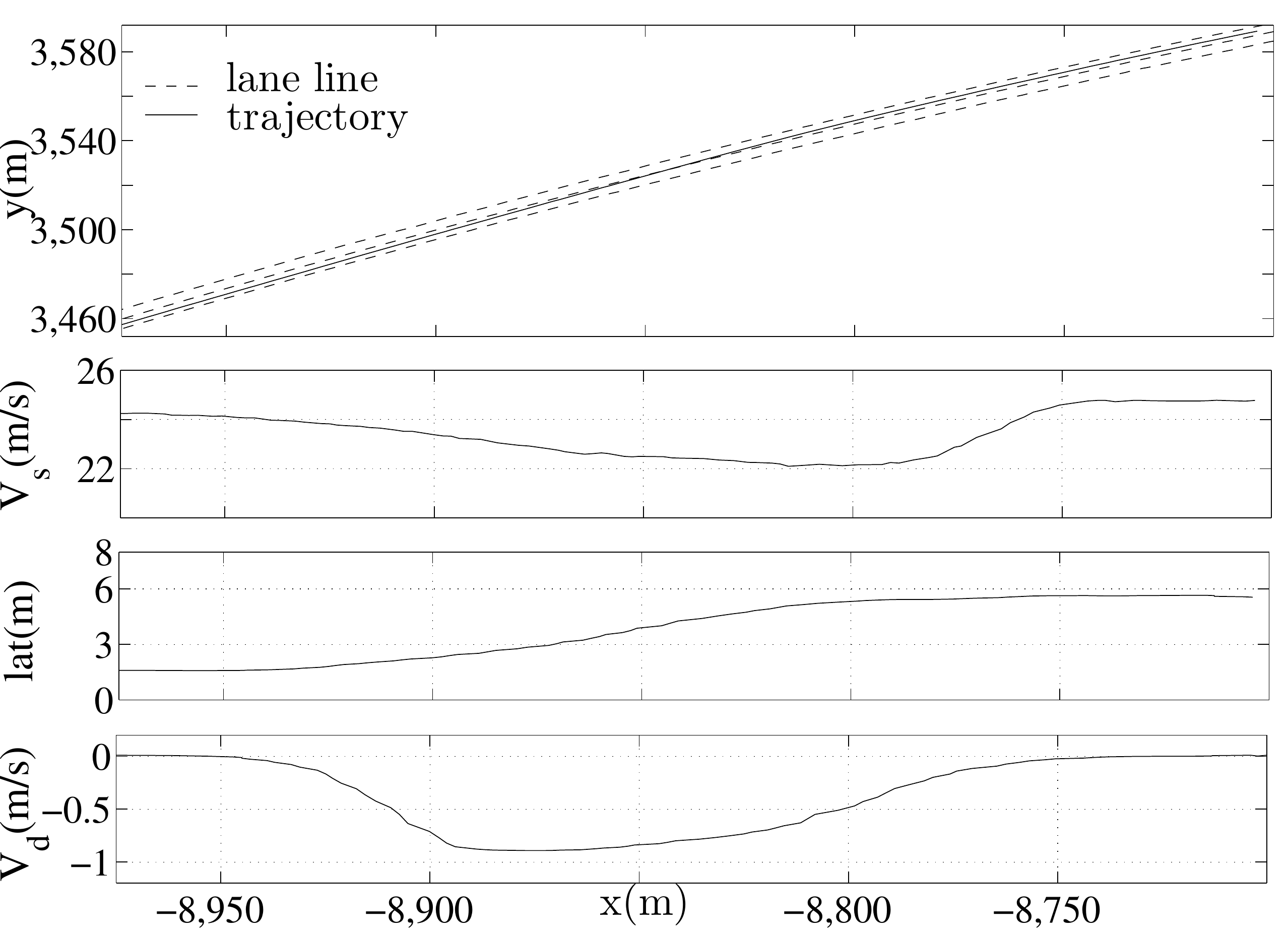}
	\caption{\myadd{States of taking over a slow vehicle. From top to bottom the plots respectively illustrate the trajectory in lanes, the longitudinal velocity, the lateral shift distance to the leftmost lane line and the lateral shift velocity to the leftmost lane line.}}
	\label{exp_lanechange1}
\end{figure}

\begin{figure}[htbp]
	\centering
	\includegraphics[width=0.95\columnwidth]{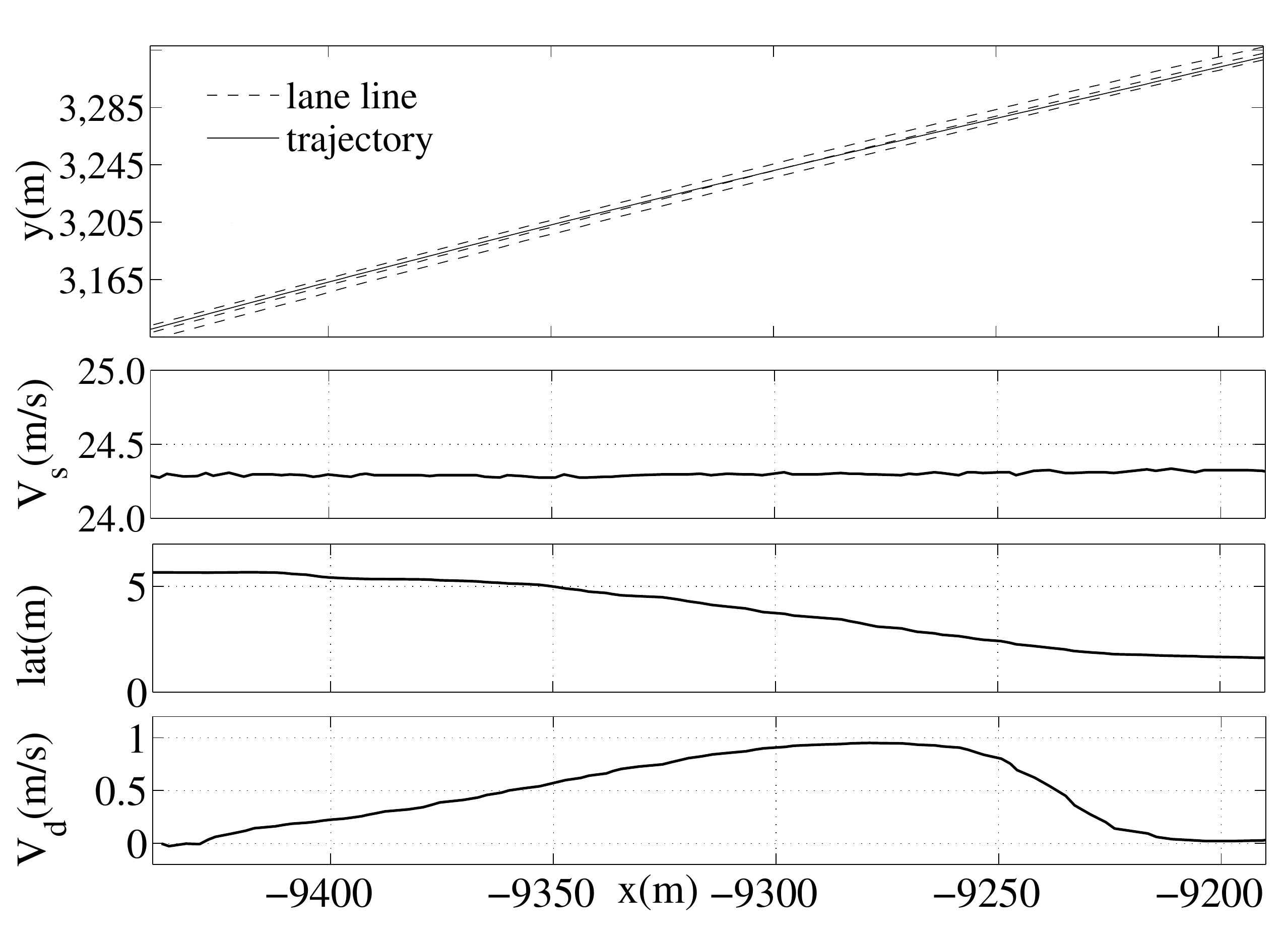}
	\caption{\myadd{States of going back to the right lane. From top to bottom the plots respectively illustrate the trajectory in lanes, the longitudinal velocity, the lateral shift distance to the leftmost lane line and the lateral shift velocity to the leftmost lane line.}}
	\label{exp_lanechange2}
\end{figure}

In Fig.\ref{exp_all} two consecutive maneuvers of keeping in the lane and changing to the left lane from overtaking a slow vehicle are illustrated. The light yellow profiles in \protect\subref{exp_lc_vt} and \protect\subref{exp_lc2_vt} are \textit{root profile} of lane keeping and \textit{end profile} of left change respectively. From top to down in Fig.\ref{exp_lanechange1} respectively depicts the global trajectory against the lane lines, the longitudinal velocity, the lateral shift to the leftmost lane line and the lateral velocity toward the leftmost line of the agent. The agent approaches the slow vehicle $ V_{56} $ without decreasing its speed until $ V_{56} $ is detected. Once $ V_{56} $ is perceived,  the agent slows down to about 22.3 m/s to keep a safe distance for lane keeping. As the maneuver of left change exists longer, the value of its cost increases. While slow $ V_{56} $ results in a decrease in the value of cost for lane keeping. So, shortly after the agent slows down, the left change maneuver is selected, along with corresponding trajectory. Fig.\ref{exp_lanechange2} illustrates the same states of the agent changing back to drive lane (the right one). Similarly, the agent keeps in the left lane for a while and selects the right change maneuver as no front vehicle on roads. As the curvy road turns to left side, the initial lateral velocity of right change is relatively faster than the one of left change. However, the overall lateral velocity is still smooth enough.

The maneuvers overtaking a slow vehicle in Fig.\ref{exp_all}, Fig.\ref{exp_lanechange1} and Fig.\ref{exp_lanechange2} suggest that the agent can maintains a safe distance to surrounding vehicles. The ability of planning reasonable maneuvers for either safe lane keeping or smooth lane change to reach target lane is validated.

\section{Conclusion}
\label{sec_conclusion}
In this paper, we proposed the SMSTP algorithm to enumerate maneuvers associated with feasible trajectories in structured environments. The SMSTP algorithm enumerates possible topological maneuvers represented by TPs which we present for the first time. It optimizes a trajectory in longitudinal and lateral direction step-wisely with heuristic information. Our optimization algorithm distinctly reduces the complexity of trajectory planning in 3D space and the computation load. It is capable of dealing with different motion models of traffic participants and lane constraints. Therefore, we can easily integrate semantic task into our SMSTP algorithm.
Besides, the algorithm provides an intuitive maneuver generation method which includes the acceleration and deceleration information in the future horizon. All non selected trajectories are stored as backups and can be selected according to high-level task and the preference of passengers.

Simulation and realistic driving experimental results show the proposed algorithm\textquotesingle s effectiveness of finding various maneuvers and generating the corresponding smooth trajectories in different traffic situations. These results motivate our future work in integrating human-machine interaction module and using machine learning methods to select a satisfying maneuver. For numerical optimization, machine learning methods are also worth investigating to replace current algorithms using empirically selected weights. \myadd{We also plan to apply the proposed algorithm into decision and planning in complicated intersection scenarios.}

\bibliographystyle{IEEEtran}
\bibliography{References}

\newpage




\end{document}